\documentclass{article}

\usepackage[final,nonatbib]{neurips_2025}

\usepackage[utf8]{inputenc} % allow utf-8 input
\usepackage[T1]{fontenc}    % use 8-bit T1 fonts
\usepackage{hyperref}       % hyperlinks
\usepackage{url}            % simple URL typesetting
\usepackage{booktabs}       % professional-quality tables
\usepackage{amsfonts}       % blackboard math symbols
\usepackage{nicefrac}       % compact symbols for 1/2, etc.
\usepackage{microtype}      % microtypography
\usepackage{xcolor}         % colors
\usepackage{upgreek}
\usepackage{titletoc}

\usepackage{amsmath}
\usepackage[numbers]{natbib}
\usepackage{amsthm}
\usepackage{cleveref}
\let\cref\Cref

\usepackage[inline]{enumitem}
\newlist{enuminline}{enumerate*}{1}
\setlist[enuminline]{label=(\roman*)}

\newcommand{\E}{\mathbb E}
\newcommand{\var}{\mathrm{var}}
\newcommand{\cov}{\mathrm{cov}}
\newcommand{\Normal}{\mathcal N}

\newcommand{\bbR}{\mathbb R}
\newcommand{\bbE}{\mathbb E}
\newcommand{\bbP}{\mathbb P}
\newcommand{\calC}{\mathcal C}

\newcommand{\calF}{\mathcal F}
\newcommand{\cv}{\mathrm{cv}}
\newcommand{\icv}{\mathrm{icv}}
\newcommand{\cvpp}{\mathrm{cv+}}
\newcommand{\pp}{\mathrm{pp}}
\newcommand{\avpp}{\mathrm{avpp}}

\newcommand{\tX}{\widetilde X}

\newcommand{\PP}{\mathrm{\texttt{PP}}}
\newcommand{\PPpp}{\mathrm{\texttt{PP+}}}

\newcommand{\loss}{\ell}
\newcommand{\iid}{i.i.d.\ }
\newcommand{\NA}{\mathrm{\tt NA}}
\newcommand{\BA}{\mathrm{\tt BA}}
\newcommand{\mNA}{\mathrm{\tt mNA}}
\newcommand{\mBA}{\mathrm{\tt mBA}}

\newcommand{\AsympCS}{AsympCS\ }

\newcommand{\prior}{\uppi}
\newcommand{\iidsim}{\overset{\mathrm{iid}}{\sim}}
\usepackage{graphicx}

\newtheorem{definition}{Definition}
\newtheorem{proposition}{Proposition}

\newtheorem{theorem}{Theorem}
\newtheorem{lemma}{Lemma}

\title{Anytime-valid, Bayes-assisted,\\Prediction-Powered Inference}

\author{%
  Valentin Kilian\thanks{Equal contribution. Order decided by coin toss.} \\
  Department of Statistics,\\
  University of Oxford\\
  \texttt{kilian@stats.ox.ac.uk} \\
   \And
  Stefano Cortinovis\footnotemark[1]\\
  Department of Statistics,\\
  University of Oxford\\
  \texttt{cortinovis@stats.ox.ac.uk} \\
  \And
  François Caron \\
  Department of Statistics,\\
  University of Oxford\\
  \texttt{caron@stats.ox.ac.uk} \\
}

\begin{document}

\maketitle

\begin{abstract}
  Given a large pool of unlabelled data and a smaller amount of labels, prediction-powered inference (PPI) leverages machine learning predictions to increase the statistical efficiency of confidence interval procedures based solely on labelled data, while preserving fixed-time validity.
  In this paper, we extend the PPI framework to the sequential setting, where labelled and unlabelled datasets grow over time. 
  Exploiting Ville's inequality and the method of mixtures, we propose prediction-powered confidence sequence procedures that are asymptotically valid uniformly over time and naturally accommodate prior knowledge on the quality of the predictions to further boost efficiency.
  We carefully illustrate the design choices behind our method and demonstrate its effectiveness in real and synthetic examples.
\end{abstract}

%%%%%%% Section 1
% !TEX root = ppics_main.tex

\section{Introduction}
\label{sec:introduction}
Increasing the sample size of an experiment is arguably the single simplest way to improve the precision of the statistical conclusions drawn from it.
However, in many fields -- such as healthcare, finance, and social sciences -- obtaining labelled data is often costly and time-consuming.
In these settings, using machine learning (ML) models to impute additional labels represents a tempting alternative to expensive data collection, albeit at the risk of introducing bias.
Prediction-powered inference (PPI)~\citep{Angelopoulos2023} is a recently introduced framework for valid statistical inference in the presence of a small labelled dataset and a large number of unlabelled examples paired with predictions from a black-box model. \looseness=-1

Formally, given an input/output pair $(X,Y)\sim \bbP=\bbP_X\times\bbP_{Y|X}$, consider the goal of estimating
\begin{align}
    \theta^\star=\underset{\theta\in\bbR}{\arg\min}~~\E[\loss_\theta(X,Y)],
    \label{eq:defthetastar1}
\end{align}
where $\loss_\theta(x,y)$ is a convex loss function parameterised by $\theta\in\bbR$.
As an example, the mean $\theta^\star=\E[Y]$ is the estimand induced by the squared loss $\loss_\theta(x,y)=(\theta - y)^2/2$.
For $t=1,2,\ldots$, we observe a sequence of independent random variables $Z_t$, either drawn from $\bbP$ (labelled sample) or from $\bbP_X$ (unlabelled sample), and we are provided with a black-box prediction rule $f$ that maps any input $x$ to a prediction $f(x)$.

Let $(X_i,Y_i)_{i\geq 1}$ and $(\widetilde X_{j})_{j\geq 1}$ denote the subsequence of labelled and unlabelled samples, respectively.
For $n=1,2,\ldots$, let $N_n$ denote the number of unlabelled samples observed before the $n$th labelled one, and assume that $N_n \geq n$, with $N_n \gg n$ in typical settings.
PPI constructs an (asymptotic) $1 - \alpha$ confidence interval (CI) $\calC_{\alpha,n}^\pp$ for $\theta^\star$, that exploits the auxiliary information encoded in $f$.
To this end, under mild assumptions, $\theta^\star$ can be expressed as the solution to \looseness=-1
\begin{align}
    g_{\theta^\star}:=\E[\loss_{\theta^\star}'(X,Y)]=0,
    \label{eq:defthetastar2}
\end{align}
where $\loss_\theta'$ is a subgradient of $\loss_\theta$ with respect to $\theta$.
The quantity $g_\theta$ in \cref{eq:defthetastar2} can be decomposed as $g_{\theta} = m_\theta + \Delta_\theta$, where
\begin{align}
    m_\theta :=\E[\loss_{\theta}'(X,f(X))] \quad\mbox{and}\quad \Delta_\theta:=\E[\loss_{\theta}'(X,Y)-\loss_{\theta}'(X,f(X))], \label{eq:m_Delta_theta}
\end{align}
where $m_\theta$ represents a measure of fit of the predictor, while $\Delta_\theta$, the \textit{rectifier}, accounts for the discrepancy between the predicted outputs $f(X)$ and the true labels $Y$. If $\calC^g_{\alpha,\theta,n}$ is a $(1-\alpha)$ confidence interval for $g_\theta$, then the PPI confidence interval $\calC_{\alpha,n}^\pp$, defined as
\begin{align}
    \calC_{\alpha,n}^\pp=\left\{\theta \mid 0\in \calC^g_{\alpha,\theta,n}\right\},
    \label{eq:ppconfidenceinterval}
\end{align}
also achieves the desired coverage, i.e.
$\Pr(\theta^\star\in \calC_{\alpha,n}^\pp)\geq 1-\alpha$. Constructing $\calC^g_{\alpha,\theta,n}$ naturally relies on estimating $g_\theta$, for which PPI defines an estimator that leverages both the unlabelled data and the prediction rule $f$.
The resulting method outperforms standard CI procedures based on the labelled data alone when $f$ is sufficiently accurate and $N_n \gg n$.
Intuitively, this is because, in this case, $\Delta_\theta$ is close to zero, while $m_\theta$ can be estimated with low variance from the unlabelled data.

Crucially, coverage of the PPI CI \eqref{eq:ppconfidenceinterval} is guaranteed only at a fixed time, i.e., for a labelled sample size $n$ fixed in advance.
This is undesirable in many practical settings -- such as online learning, real-time monitoring, or sequential decision-making -- where it is essential to continuously draw conclusions as new data arrive.
In this work, we address this by proposing an \textit{anytime-valid} extension of the PPI CI \eqref{eq:ppconfidenceinterval}.
That is, we define a confidence sequence $(\calC_{\alpha,n}^\avpp)_{n\geq 1}$, satisfying asymptotically the stronger coverage guarantee
\[
    \Pr(\theta^\star\in \calC_{\alpha,n}^\avpp \text{ for all }n\geq 1)\geq 1-\alpha,
\]
while still taking advantage of the prediction rule $f$. Analogously to standard PPI, we construct a confidence sequence $(\calC^g_{\alpha,\theta,n})_{n \geq 1}$ for $g_\theta$ and define $\calC_{\alpha,n}^\avpp$ through \cref{eq:ppconfidenceinterval} for $n \geq 1$.
While our approach is agnostic to the specific form of the confidence sequence $(\calC^g_{\alpha,\theta,n})_{n \geq 1}$, we mainly focus on asymptotic confidence sequences~\citep{WaudbySmith2024a}, as they provide a versatile time-uniform analogue of standard CLT-based CIs that applies to the PPI framework above in full generality.
Moreover, being based on the method of mixtures~\citep{Ville1939,Robbins1970,Lai1976}, they can readily accommodate prior information on the quality of the prediction model $f$.
In particular, by means of a zero-centred prior on the rectifier $\Delta_\theta$, we obtain tighter confidence sequences when the predictions are good, extending the fixed-time Bayes-assisted approach of \citet{Cortinovis2025}.

The remainder of the paper is organised as follows.
\Cref{sec:relatedwork} reviews related work.
\Cref{sec:asympCS} provides background on (asymptotic) confidence sequences and discusses how prior information may be incorporated into their construction.
\Cref{sec:ppi} presents PPI in the context of control-variate estimators, whose asymptotic properties are crucial for our approach to anytime-valid, Bayes-assisted PPI, which is described in \cref{sec:abppi}.
\Cref{sec:experiments} demonstrates the benefits of our method on synthetic and real data.
Finally, \cref{sec:discussion} discusses limitations and further extensions of our approach. Proofs and additional experiments are provided in the Supplementary Material.
%%%%%%% Section 2
% !TEX root = ppics_main.tex
\section{Related Work}
\label{sec:relatedwork}

PPI was introduced by \citet{Angelopoulos2023} as a general framework for valid statistical inference with black-box machine learning predictors, and was later extended in \citet{Angelopoulos2023a}.
Closely related ideas appear in the literatures on semi-supervised inference, missing-data methods, survey sampling, and double machine learning \citep{Robins1995,Saerndal2003,Chernozhukov2018,Zhang2019,Zhang2022}.
More recently, \citet{Cortinovis2025} proposed a Bayes-assisted variant of PPI. All of these contributions target fixed-time confidence intervals. \looseness=-1

Confidence sequences were first introduced by \citet{Darling1967} and developed further by \citet{Robbins1970} and \citet{Lai1976}, building on earlier work by \citet{Ville1939} and \citet{Wald1945}. 
Interest has surged again in recent years \citep{WaudbySmith2024,WaudbySmith2024a}, motivated by applications such as A/B testing.
The notion is closely linked to e-values~\citep{Ramdas2023,Ramdas2024}.
Building on the e-value framework, and on earlier work by \citet{Zrnic2024} and \citet{WaudbySmith2024}, \citet{Csillag2025} proposed an exact, time-uniform PPI method that yields confidence sequences under stronger conditions (e.g., existence of bounded e-values) and does not leverage prior knowledge about the ML prediction quality.
Furthermore, application of their method requires an active-learning setup in which, for each $t$, the observation $Z_t$ can be labelled with strictly positive probability.
In particular, it is not applicable to deterministic sequences of observations, such as those describing a large initial pool of unlabelled data followed by a stream of labelled data -- the main focus of our experiments.

In the setting of double machine learning and semiparametric inference, \citet{Dalal2024} and \citet{WaudbySmith2024a} derive asymptotic confidence sequences for target parameters in the presence of high-dimensional nuisance components.
%%%%%%% Section 3
% !TEX root = ppics_main.tex

\section{Asymptotic (Bayes-assisted) confidence sequences}
\label{sec:asympCS}

In this section, we begin with background on (asymptotic) confidence sequences (CS).
We then show how prior information can be incorporated into asymptotic CS procedures, leading to asymptotic Bayes-assisted confidence sequences. 

\subsection{Background}\label{sec:background}
We start by defining an exact confidence sequence~\citep{Darling1967}, a time-uniform analogue of classical CIs.
% We start with the definition of a confidence sequence~\citep{Darling1967}, which is the time-uniform analogue of a classical fixed-sample confidence interval.
\begin{definition}[Confidence sequence]
\label{def:CS}
Let $(\calC_{\alpha,t})_{t\geq 1}$ be a sequence of random subsets of $\bbR$. For $\alpha\in(0,1)$, $(\calC_{\alpha,t})_{t\geq 1}$ is a $1-\alpha$ confidence sequence for a fixed parameter $\mu\in\bbR$ if
\begin{equation}
\Pr(\mu\in \calC_{\alpha,t}\text{  ~for all }t\geq 1)\geq 1-\alpha.
\label{eq:coverageCS}
\end{equation}
\end{definition}
We now introduce the notion of an asymptotic confidence sequence (AsympCS) \citep{WaudbySmith2024a,Dalal2024}.
\begin{definition}[Asymptotic confidence sequence]
\label{def:asympCS}
Let $\alpha\in(0,1)$ and $(a_t)_{t\geq 1}$ be a real sequence such that $\lim_{t\to\infty} a_t=0$. Let $(\widehat\mu_t)_{t\geq 1}$ be a consistent sequence of estimators of $\mu$. The sequence of random intervals $(\calC_{\alpha,t})_{t\geq 1}$, with $\calC_{\alpha,t}=[\widehat\mu_t-L_t,\widehat\mu_t+U_t]$ and $L_t>0$, $U_t>0$, is said to be an \textit{asymptotic confidence sequence with (little-o) approximation rate $a_t$} if there exists a (usually unknown) confidence sequence $(\calC^\star_{\alpha,t})_{t\geq 1}$, with $\calC^\star_{\alpha,t}=[\widehat\mu_t-L_t^\star,\widehat\mu_t+U_t^\star]$, such that
\[\Pr(\mu\in \calC^\star_{\alpha,t}\text{ for all }t\geq 1)\geq 1-\alpha\]
and, almost surely as $t\to\infty$, $\max\{L_t^\star-L_t,U_t^\star-U_t\} = o(a_t)$.
\end{definition}
Thus, an asymptotic CS may be regarded as an approximation of an exact CS that becomes arbitrarily accurate in the limit.
It is worth noting that, while classical fixed-sample asymptotic CIs rely on \textit{convergence in distribution} of the scaled estimators, asymptotic confidence sequences rely on the \textit{almost sure convergence at a given rate} of the centred lower and upper bounds relative to those of an underlying exact CS.
The following is an example of an asymptotic CS that applies to \iid data.
\begin{theorem}
Let $(Y_t)_{t\geq 1}$ be a sequence of \iid random variables with mean $\mu$ and such that $\E\vert Y_1\vert^{2+\delta}<\infty$ for some $\delta>0$. For any $t\geq 1$, let $\overline Y_t$ be the sample mean, and $\widehat\sigma_t^2$ be the sample variance based on the first $t$ observations. For any parameter $\rho>0$, the sequence of intervals defined as \looseness=-1
\begin{equation}
\calC^\NA_{\alpha,t}(\overline Y_t,\widehat\sigma_t;\rho):=\left[ \overline Y_t \pm \frac{\widehat\sigma_t}{\sqrt{t}} \sqrt{\left(1+\frac{1}{t\rho^2}\right)\log\left(\frac{t\rho^2+1}{\alpha^2}\right)} \right ]
\label{eq:NonBayesianConfidenceSequence}
\end{equation}
forms a $(1-\alpha)$--\AsympCS with approximation rate $1/\sqrt{t\log{t}}$ for $\mu$.
\label{thm:AsympCSlocalprioriid}
\end{theorem}
For the sequel, it is useful to highlight some aspects of the proof of this theorem.
First, if the random variables $(Y_t)_{t\geq 1}$ were Gaussian with variance $\sigma^2$, then $\calC^\NA_{\alpha,t}(\overline Y_t,\sigma;\rho)$ would be an exact CS.
This follows from combining the method of mixtures for nonnegative martingales with Ville's inequality~\cite{Ville1939,Robbins1970,Lai1976,Howard2021}.
Second, the proof relies on KMT strong coupling \citep{Komlos1975,Major1976}: there exists \iid Gaussian random variables $(W_t)_{t\geq 1}$ with mean $\mu$ and variance $\var(Y)$ such that 
\[
    \frac{1}{t}\sum_{i=1}^t Y_i =\frac{1}{t}\sum_{i=1}^t W_i +o\left(\frac{1}{\sqrt{t \log t}} \right)\text{ a.s. as }t\to\infty.
\]
Such a coupling plays a central role in constructing asymptotic confidence sequences, serving as a substitute for the CLT assumption underlying classical fixed-sample CIs.
The construction in \cref{thm:AsympCSlocalprioriid} extends beyond the \iid case, provided a similar coupling exists.

\begin{theorem}
Let $(\widehat\mu_t)_{t\geq 1}$ be a consistent sequence of estimators of $\mu$. Assume that there exists a sequence of \iid Gaussian random variables $(W_i)_{i\geq 1}$, with mean $\mu$ and variance $\sigma^2$, such that
\begin{equation}
\label{condition1}
\widehat{\mu}_t=\frac{1}{t} \sum_{i=1}^t W_i+o\left(\frac{1}{\sqrt{t \log t}}\right)\text{ a.s. as }t\to\infty.
\end{equation}
Let $(\widehat\sigma^2_{t})_{t\geq 1}$ be a consistent sequence of estimators of $\sigma^2$ with $\vert\widehat\sigma_t-\sigma\vert=o\left(\frac{1}{\log t}\right)$ a.s. Then, for any parameter $\rho>0$, the sequence of intervals $(\calC^\NA_{\alpha,t}(\widehat\mu_t,\widehat\sigma_t;\rho))_{t\geq 1}$ forms a $(1-\alpha)$--\AsympCS with approximation rate $1/\sqrt{t\log{t}}$ for $\mu$. \label{thm:AsympCSlocalpriornoniid}
\end{theorem}

The asymptotic CS \eqref{eq:NonBayesianConfidenceSequence} includes a tuning parameter $\rho$, which can be chosen so as to minimise the width of the interval at a specified time $t$; see \cite[Appendix~B.2]{WaudbySmith2024a}.
However, this method does not allow the incorporation of prior information about the parameter of interest to yield tighter intervals when the data align with those assumptions: the width of \cref{eq:NonBayesianConfidenceSequence} is indeed independent of $\overline Y_t$.

\subsection{Asymptotic Bayes-assisted confidence sequences}
\label{sec:asympbacs}
To address this, we introduce a Bayes-assisted analogue of \cref{thm:AsympCSlocalprioriid}. 
\begin{theorem}[Bayes-assisted AsympCS -- \iid case]
Let $(Y_t)_{t\geq 1}$ be a sequence of \iid random variables with unknown mean $\mu$ and unknown variance $\sigma^2$, and such that $\E\vert Y_1\vert^{2+\delta}<\infty$ for some $\delta>0$. For any $t\geq 1$, let $\overline Y_t$ be the sample mean, and $\widehat\sigma_t^2$ be the sample variance based on the first $t$ observations. Let $\eta_t:\bbR\to(0,\sqrt{t/(2\pi)})$ be defined as
\begin{align}
\eta_t(z)=\int_{-\infty}^\infty \Normal\left(z;\zeta , 1/t\right)\prior(\zeta)d\zeta.
\label{eq:marginaldensity}
\end{align}
where $\prior$ is a continuous and proper prior density on $\bbR$, strictly positive in a neighbourhood of $\mu/\sigma$. Then
\begin{align}
\calC^\BA_{\alpha,t}(\overline Y_t,\widehat\sigma_t;\prior):=\left[ \overline Y_t \pm \frac{\widehat\sigma_t}{\sqrt t} \sqrt{\log\left(\frac{t}{2\pi\alpha^2\eta_t(\overline Y_t/\widehat\sigma_t)^2}\right)}  \right]
\label{eq:BayesianConfidenceSequence}
\end{align}
forms a $(1-\alpha)$--\AsympCS with approximation rate $1/\sqrt{t\log{t}}$ for $\mu$.
\label{thm:AsympCSglobalprioriid}
\end{theorem}
In \cref{thm:AsympCSglobalprioriid}, the density $\prior$ encodes prior beliefs about the ratio $\mu/\sigma$.
Under this prior, $\eta_t$ represents the marginal density of the standardised mean $\overline Y_t/\sigma$ that would arise if the observations $(Y_t)_{t \geq 1}$ were normally distributed.
In contrast to the non-assisted \AsympCS \eqref{eq:NonBayesianConfidenceSequence}, the width of the Bayes-assisted \AsympCS \eqref{eq:BayesianConfidenceSequence} varies with $\overline Y_t/\widehat\sigma_t$: when the data align with the prior, $\eta_t(\overline Y_t/\widehat\sigma_t)$ is large and the interval narrows; when they conflict, $\eta_t(\overline Y_t/\widehat\sigma_t)$ is small and the interval widens.
It is worth emphasising that, even when the prior is strongly misspecified, the Bayes-assisted \AsympCS \eqref{eq:BayesianConfidenceSequence} remains valid. \looseness=-1
In the case of a Gaussian prior $\prior$ centred at $\mu_0$ with variance $\tau^2$, we obtain the following \AsympCS:
\begin{equation}
    \calC^\BA_{\alpha,t}(\overline Y_t,\widehat\sigma_t;\Normal(\cdot;\mu_0,\tau^2))=\left[ \overline Y_t \pm \frac{\widehat\sigma_t}{\sqrt t} \sqrt{\log\left(\frac{t\tau^2 +1}{\alpha^2} \right) + \frac{(\overline Y_t/\widehat\sigma_t -\mu_0)^2}{\tau^2+1/t}  }  \right].
    \label{eq:GaussianConfidenceSequence}
\end{equation}
Setting $\rho = \tau$ allows a direct comparison between \eqref{eq:GaussianConfidenceSequence} and its non-assisted counterpart \eqref{eq:NonBayesianConfidenceSequence}.
When the data agree with the prior -- i.e., $\overline Y_t/\widehat\sigma_t -\mu_0\simeq 0$ -- the Bayes-assisted interval is narrower than the non-assisted one. 
Conversely, if the data conflict with the prior, $(\overline Y_t/\widehat\sigma_t -\mu_0)^2$ is large and the Bayes-assisted \AsympCS becomes wider than \eqref{eq:NonBayesianConfidenceSequence}.

The proof of \cref{thm:AsympCSglobalprioriid} is similar to that of \cite[Theorem~2.2]{WaudbySmith2024a}.
First, note that $\calC^\BA_{\alpha,t}(\overline Y_t,\var(Y);\prior)$ would be an exact CS if the observations were normally distributed.
This follows from an application of the method of mixtures for nonnegative martingales, using the prior $\prior$ as mixing density, together with Ville's inequality. Second, we use KMT strong coupling to approximate in an almost sure sense $\overline Y_t$ by a sample average of \iid Gaussian random variables. As in the non-assisted case, \cref{thm:AsympCSglobalprioriid} can be extended to the non-\iid setting, as long as one can find such a strong coupling.
\begin{theorem}[Asymptotic Bayes-assisted CS -- non-\iid case]
Consider the same notation and assumptions as in \cref{thm:AsympCSlocalpriornoniid}.
Let $\prior$ be a continuous and proper prior density on $\bbR$, strictly positive in a neighbourhood of $\mu/\sigma$, and let $\eta_t$ be the density \eqref{eq:marginaldensity} for any $t\geq 1$.
Then, the sequence of intervals $(\calC^\BA_{\alpha,t}(\widehat\mu_t,\widehat\sigma_t;\prior))_{t\geq 1}$ forms a $(1-\alpha)$--\AsympCS with approximation rate $1/\sqrt{t\log{t}}$ for $\mu$.

\label{thm:AsympCSglobalpriornoniid}
\end{theorem}

\subsection{Asymptotic Type-I error control}

The asymptotic confidence sequences defined above satisfy an asymptotic version of time-uniform Type-I error control (in the sense of \cite[§2.5]{WaudbySmith2024a}; see also \cite{Bibaut2022}).
\begin{theorem}[Asymptotic Type-I error control]
\label{thm:typeIerror}
Assume the hypotheses of one of \cref{thm:AsympCSlocalprioriid,thm:AsympCSlocalpriornoniid,thm:AsympCSglobalprioriid,thm:AsympCSglobalpriornoniid}, and let $(\calC_{\alpha,t})$ be the corresponding $(1-\alpha)$--\AsympCS for $\mu$. Then  
\begin{align}
\underset{m \rightarrow \infty}{\lim\inf} \Pr\left(\mu \in \calC_{\alpha,t}\text{ ~for all }t \geq m\right) \geq 1-\alpha.
\end{align}
\end{theorem}\looseness=-1
%%%%%%% Section 4
% !TEX root = ppics_main.tex

\section{Control variates and PPI: background and strong coupling}
\label{sec:ppi}

Prediction-powered inference (PPI) closely relates to control variates, a standard variance-reduction method in Monte Carlo estimation \citep[\S4.1]{Glasserman2003}.
In fact, each PPI estimator can be expressed as a control-variate estimator.
We begin with a review of control variates and derive a KMT-type strong-coupling result for these estimators, and then provide additional background on PPI.

\subsection{Control variates: definitions and KMT strong coupling}
\label{subsect:control}
Let $(U, V)$ be real-valued random variables with finite variance, and consider the goal of estimating $\gamma=\E[V]$ from an \iid sample ${(U_i, V_i)}_{i=1}^n$.
If $\mu = \E[U]$ is known, the control-variate estimator (CVE) of $\gamma$ is defined as
\begin{align}
    \widehat \gamma^{\icv}_\lambda=\overline V - \lambda (\overline U - \mu)=\frac{1}{n}\sum_{i=1}^n \left( V_i - \lambda(U_i-\mu)\right), \label{eq:cvestimator}
\end{align}
where $\overline U$ and $\overline V$ denote the empirical means of $(U_i)_{i=1}^n$ and $(V_i)_{i=1}^n$, respectively, $\lambda\in\bbR$ is a tunable coefficient, and the term $U_i-\mu$ acts as a control variate.
The estimator $\widehat{\gamma}^{\icv}_\lambda$ is unbiased, consistent, and has variance $\var(\widehat \gamma^{\icv}_\lambda)=(\var(V)-2\lambda \cov(U,V)+\lambda^2 \var(U))/n$.
Compared to the standard sample mean estimator $\overline{V}$, which attains variance $\var(\overline{V}) = \var(V)/n$, using $\widehat{\gamma}^{\icv}_\lambda$ results in variance reduction when $\lambda < 2\cov(U, V)/\var(U)$.
The minimum variance is achieved at the optimal coefficient $\lambda^\star = \cov(U,V) / \var(U)$, for which $\var(\widehat \gamma^{\icv}_{\lambda^\star}) =(1-\rho_{U,V}^2)\var(\overline V)$, where $\rho_{U,V}$ is the correlation between $U$ and $V$.
That is, stronger correlation leads to greater variance reduction.

In practice, both $\mu$ and $\lambda^\star$ are typically unknown.
When this is the case, given an additional \iid sample $(\widetilde U_j)_{j=1}^{N_n}$, independent of ${(U_i, V_i)}_{i=1}^n$, where $\widetilde U_1$ has the same distribution as $U$, one can estimate $\mu$ by $\widehat \mu=\frac{1}{N_n} \sum_{j=1}^{N_n}\widetilde U_j $ and plug it into \cref{eq:cvestimator}. For fixed $\lambda$, this gives
\begin{align}
    \widehat \gamma_\lambda^{\cv}=\overline V - \lambda (\overline U - \widehat\mu)=\frac{1}{n}\sum_{i=1}^n \left( V_i - \lambda(U_i-\widehat\mu)\right). \label{eq:cvestimator2}
\end{align}
Similarly, $\lambda^\star$ may be estimated from data as $\widehat \lambda = \widehat\cov((U_i,V_i)_{i=1}^n)/\widehat\var((U_i)_{i=1}^n)$,

where $\widehat\var(\cdot)$ and $\widehat\cov(\cdot)$ denote the sample variance and covariance, respectively. Plugging $\widehat \lambda$ into \eqref{eq:cvestimator2} defines $\widehat \gamma^{\cvpp} := \widehat \gamma_{\widehat\lambda}^{\cv}$, which is similar to the semi-supervised least squares estimator of \citet[Eq.~(2.15)]{Zhang2019}. As discussed in \cref{sec:asympCS}, deriving an \AsympCS requires a strong coupling between the estimator and a sequence of \iid Gaussian random variables. We now establish this coupling, a key ingredient for \AsympCS for CVEs (and, in particular, for PPI estimators).

\begin{proposition}[Asymptotics for CVEs]
\label{prop:asympcvestimator}
 Assume $\E\vert U\vert^{2+\delta}$ and $\E\vert V\vert^{2+\delta}<\infty$ for some $0<\delta<1$. Then, almost surely as $n\to\infty$,
    \begin{align}
        \widehat \gamma^{\cvpp}&=\widehat \gamma^{\cv}_{\lambda^\star}+o\left(\frac{1}{\sqrt{n\log n}}\right)=\overline V - \lambda^\star (\overline U - \widehat \mu)+o\left(\frac{1}{\sqrt{n\log n}}\right).\label{eq:strongcvvscvpp}
    \end{align}
\end{proposition}

\begin{proposition}[KMT coupling for CVEs]
    \label{prop:strassencvestimator}
    Assume $\mathbb{E}\vert U\vert^{2+\delta}$ and $\mathbb{E}\vert V\vert^{2+\delta}<\infty$ for some $0<\delta<1$. Assume additionally that $|\frac{n}{N_n}-r|=O(1/n^{1-a})$ with $0<a<2/(2+\delta)$, for some $r\in[0,1]$. Then, there exist \iid Gaussian random variables $(W^{\cv}_i)_{i\geq 1}$ with mean $\gamma$ and variance
    $$
        \nu_\lambda^{\cv} := \var(V - \lambda U )+r\var(\lambda U)=\var(V)-2\lambda\cov(U,V)+\lambda^2 (1+r)\var(U)
    $$
    such that, almost surely as $n\to\infty$,
    \begin{align}
    \widehat \gamma_\lambda^{\cv}=\frac{1}{n}\sum_{i=1}^n W^{\cv}_i +o\left(\frac{1}{\sqrt{n\log n}}\right).
    \label{eq:CVEgammacoupling}
    \end{align}
    Similarly, there exist \iid Gaussian random variables $(W^{\cvpp}_i)_{i\geq 1}$ with mean $\gamma$ and variance
    $
        \nu^{\cvpp} :=\nu_{\lambda^\star}^{\cv} =\var(V)\left[1-(1-r)\rho_{U,V}^2\right]
    $
    such that, almost surely as $n\to\infty$,
    \begin{align}
    \widehat \gamma^{\cvpp}=\frac{1}{n}\sum_{i=1}^n W^{\cvpp}_i +o\left(\frac{1}{\sqrt{n\log n}}\right).
    \label{eq:CVEgammacoupling2}
    \end{align}
    The estimators
    \begin{align}
        \widehat \nu_\lambda^{\cv}((U_i,V_i)_{i=1}^n,(\widetilde U_j)_{j=1}^{N_n})&=\frac{1}{n-2}\sum_{i=1}^n (V_i - \overline V-  \lambda (U_i-\overline U))^2+\frac{n\lambda^2}{N_n(N_n-1)}\sum_{j=1}^{N_n} (\widetilde U_j-\widehat\mu)^2\label{eq:nuestcv}\\
        \widehat \nu^{\cvpp}((U_i,V_i)_{i=1}^n)&=\frac{1-n/N_n}{n-2}\sum_{i=1}^n (V_i - \overline V-  \widehat\lambda (U_i-\overline U))^2+\frac{n/N_n}{n-1}\sum_{i=1}^n (V_i-\overline V)^2\label{eq:nuestcvpp}
    \end{align}
    are consistent estimators of $\nu_\lambda^\cv$ and $\nu^\cvpp$, respectively, where $\widehat \mu=\frac{1}{N_n} \sum_{j=1}^{N_n}\widetilde U_j$.

\end{proposition}

\subsection{PPI estimators: definitions and asymptotic properties}

Owing to \cref{eq:defthetastar2}, the PPI estimator $\widehat \theta_n$ is the value of $\theta$ that solves the equation $\widehat g_{\theta,n}=0$, where $\widehat g_{\theta,n}= \widehat m_{\theta,n} + \widehat \Delta_{\theta,n}$ is an estimator of $g_\theta$. Here, $\widehat m_{\theta,n}$ and $\widehat \Delta_{\theta,n}$ are estimators of $m_\theta$ and $\Delta_\theta$, respectively.
A typical choice for $\widehat m_{\theta,n}$ is the sample mean of the unlabelled data,
\begin{align}
    \widehat m_{\theta,n}&=\frac{1}{N_n}\sum_{j=1}^{N_n} \loss_{\theta}'(\widetilde X_j,f(\widetilde X_j)). \label{eq:msamplemean}
\end{align}
Different choices for $\widehat \Delta_{\theta,n}$ have been proposed in the literature, leading to different PPI estimators.
\paragraph{Standard PPI.}
\citet{Angelopoulos2023} use the sample mean
\begin{align}
    \widehat \Delta^\PP_{\theta,n}&=\frac{1}{n}\sum_{i=1}^{n} \left(\loss_{\theta}'(X_i,Y_i)-\loss_{\theta}'(X_i,f(X_i))\right)\label{eq:deltasamplemean}
\end{align}
as an estimator for $\Delta_\theta$. Combining \cref{eq:deltasamplemean} with \cref{eq:msamplemean},
\begin{align}
    \widehat g^\PP_{\theta,n} =\widehat m_{\theta,n}+\widehat \Delta^\PP_{\theta,n}= \left[\frac{1}{n}\sum_{i=1}^{n} \loss_{\theta}'(X_i,Y_i)\right]-\left(\left[\frac{1}{n}\sum_{i=1}^n \loss_{\theta}'(X_i,f(X_i))\right]- \widehat m_{\theta,n}  \right )
    \label{eq:ghatppi}
\end{align}
is a CVE, with control variate $\loss_{\theta}'(X_i,f(X_i))-\widehat m_{\theta,n}$ and control-variate parameter $\lambda=1$.
For the squared loss, the estimator $\widehat\theta_n^\PP$ solving $\widehat g^\PP_{\theta,n} = 0$ also takes the control-variate form
\begin{align}
    \widehat\theta_n^\PP= \frac{1}{n}\sum_{i=1}^{n}Y_i - \left( \frac{1}{n}\sum_{i=1}^{n} f(X_i) - \frac{1}{N_n}\sum_{j=1}^{N_n} f(\tX_j)\right), \label{eq:thetahatppi}
\end{align}
with control variate $f(X_i)-\frac{1}{N_n}\sum_{j=1}^{N_n} f(\tX_j)$ and $\lambda=1$.
\paragraph{PPI\texttt{++}.}
\citet{Angelopoulos2023a} extend the standard PPI estimator \eqref{eq:ghatppi} by allowing the control-variate parameter $\lambda$, which they call \textit{power-tuning} parameter, to take values other than $1$. The resulting estimator is
\begin{align}
   \widehat \Delta^\PPpp_{\theta,n}=\widehat \Delta^\PP_{\theta,n} - (\widehat \lambda_{\theta,n} -1 )\left( \frac{1}{n}\left [\sum_{i=1}^{n} \loss_{\theta}'(X_i,f(X_i))\right ]  -\widehat m_{\theta,n}\right), \label{eq:deltahatpowertuning}
\end{align}
where $\widehat\lambda_{\theta,n}$ is the estimator $\widehat\lambda_{\theta,n} = \widehat\cov\left((\loss_{\theta}'(X_i,Y_i),\loss_{\theta}'(X_i,f(X_i)))_{i=1}^{n}\right) / \widehat\var\left((\loss_{\theta}'(X_i,f(X_i)))_{i=1}^{n}\right)$. In this case, $\widehat \Delta^\PPpp_{\theta,n}$ is a CVE with centred control variate $\loss_{\theta}'(X_i,f(X_i))-\widehat m_{\theta,n}$, which depends only on the black-box predictions. As a result of this choice,
\begin{align}
    \widehat g^\PPpp_{\theta,n} =\widehat m_{\theta,n}+\widehat \Delta^\PPpp_{\theta,n}= \left[\frac{1}{n}\sum_{i=1}^{n} \loss_{\theta}'(X_i,Y_i)\right]-  \widehat\lambda_{\theta,n}\left(\left[\frac{1}{n}\sum_{i=1}^n \loss_{\theta}'(X_i,f(X_i))\right]- \widehat m_{\theta,n}  \right )
    \label{eq:ghatppipp}
\end{align}
is also a CVE. Under the squared loss, we obtain
\begin{align}
    \widehat\theta_n^\PPpp = \frac{1}{n}\sum_{i=1}^{n}Y_i - \widehat\lambda_{0,n}\left( \frac{1}{n}\sum_{i=1}^{n} f(X_i) - \frac{1}{N_n}\sum_{j=1}^{N_n} f(\tX_j)\right),
    \label{eq:thetahatppipp}
\end{align}
where in this case $\widehat\lambda_{\theta,n}=\widehat\lambda_{0,n}$ for all $\theta$. 

Standard asymptotic confidence intervals for PPI and PPI\texttt{++} rely on CLTs for the estimators  $\widehat g_{\theta,n}$, $\widehat m_{\theta,n}$ and $\widehat \Delta_{\theta,n}$.
In contrast, constructing asymptotic confidence sequences requires almost sure approximations by averages of \iid Gaussian variables. Since the estimators for $g_\theta$, $m_\theta$ and $\Delta_\theta$ are all CVEs, the asymptotic results of \cref{prop:asympcvestimator} and the KMT coupling of \cref{prop:strassencvestimator} both apply.
%%%%%%% Section 5
% !TEX root = ppics_main.tex
\section{Anytime-valid, Bayes-assisted, prediction-powered inference}
\label{sec:abppi}
In this section, we show how the results of \cref{sec:asympCS,sec:ppi} can be combined in the context of PPI to obtain \AsympCS for $g_\theta$.
For any $\theta\in\bbR$ and $i\geq1$, let $U_{\theta,i}=\loss_{\theta}'(X_i,f(X_i))$, $\widetilde U_{\theta,i}=\loss_{\theta}'(\widetilde X_i,f(\widetilde X_i))$ and $V_{\theta,i}=\loss_{\theta}'(X_i,Y_i)$.
Define $\overline V_{\theta,n}=\frac{1}{n}\sum_{i=1}^n V_{\theta,i}$ and $\overline U_{\theta,n}=\frac{1}{n}\sum_{i=1}^n U_{\theta,i}$. In the following, we assume $\E\vert U_{\theta,i}\vert^{2+\delta}$, $\E\vert \widetilde U_{\theta,i}\vert^{2+\delta}$ and $\E\vert V_{\theta,i}\vert^{2+\delta}<\infty$ for some $0<\delta<1$, and that $|\frac{n}{N_n}-r|=O(1/n^{1-a})$ with $0<a<2/(2+\delta)$ for some $r\in[0,1]$.
\subsection{Anytime-valid PPI}
\label{sec:avppi}
We first derive \AsympCS that do not incorporate prior information on the accuracy of the black-box predictor.
The following result follows directly from \cref{prop:strassencvestimator} and \cref{thm:AsympCSlocalpriornoniid}, owing to the control-variate form of the PPI estimator $\widehat g^\PP_{\theta,n}$ \eqref{eq:ghatppi} and the PPI\texttt{++} estimator $\widehat g^\PPpp_{\theta,n}$ \eqref{eq:ghatppipp}.
\begin{proposition}
  \label{prop:avppi_Cg}
  Let $\widehat g_{\theta,n}$ be either the PPI \eqref{eq:ghatppi} or the PPI\texttt{++}~\eqref{eq:ghatppipp} estimator.
  For PPI, let $(\widehat\sigma^{g}_{\theta,n})^2=\widehat\nu_1^\cv((U_{\theta,i},V_{\theta,i})_{i=1}^n,(\widetilde U_{\theta,j})_{j=1}^{N_n})$ (see \eqref{eq:nuestcv}).
  For PPI\texttt{++}, let $(\widehat\sigma^{g}_{\theta,n})^2=\widehat\nu^\cvpp((U_{\theta,i},V_{\theta,i})_{i=1}^n)$ (see \eqref{eq:nuestcvpp}).
  Then, for any $\rho > 0$, the sequence of intervals defined as
  $\calC^{g}_{\alpha,\theta,n}=\calC^\NA_{\alpha,n}(\widehat g_{\theta,n},\widehat\sigma^g_{\theta,n};\rho)$ forms a $(1-\alpha)$--\AsympCS with approximation rate $1/\sqrt{n\log{n}}$ for $g_\theta$ and asymptotic Type-I error control. \looseness=-1
\end{proposition}
\subsection{Anytime-valid, Bayes-assisted, PPI}
\label{sec:avbappi}
In many modern applications, extremely accurate black-box predictors are available (e.g., \cite{CreditCard2015,Graphcast2023,AlphaFold2021}).
When this is the case, we can leverage this prior information to obtain tighter AsympCS for $g_\theta$ via a zero-mean prior on $\Delta_\theta$.
Following the decomposition in \cref{eq:m_Delta_theta}, we combine an AsympCS for $m_\theta$ (\cref{thm:m}) with a Bayes-assisted AsympCS for $\Delta_\theta$ (\cref{thm:delta}). 
\begin{proposition}[AsympCS for $m_\theta$]
\label{thm:m}
Let $\widehat m_{\theta,n}$ and $(\widehat \sigma^f_{\theta,n})^2$ be the sample mean \eqref{eq:msamplemean} and sample variance of $(\loss_{\theta}'(\widetilde X_j,f(\widetilde X_j)))_{j=1}^{N_n}$. Let $\delta\in(0,1)$. For any $\rho>0$, $\mathcal R_{\delta,\theta,n}=\calC^\NA_{\delta,n}(\widehat m_{\theta,n},\widehat\sigma^f_{\theta,n};\rho)$ forms a $(1-\delta)$--\AsympCS with approximation rate $1/\sqrt{n\log{n}}$ for $m_\theta$ and asymptotic Type-I error control.
\end{proposition}
\begin{proposition}[Bayes-assisted \AsympCS for $\Delta_\theta$]
\label{thm:delta}
For PPI, let $\widehat\Delta_{\theta,n}$ and $(\widehat \sigma^\Delta_{\theta,n})^2$ be the sample mean \eqref{eq:deltasamplemean} and sample variance of $(V_{\theta,i}-U_{\theta,i})_{i=1}^n$.
For PPI\texttt{++}, let $\widehat\Delta_{\theta,n}$ be the control-variate estimator \eqref{eq:deltahatpowertuning} and $(\widehat \sigma^\Delta_{\theta,n})^2=\widehat\nu^\cvpp((U_{\theta,i},V_{\theta,i}-U_{\theta,i})_{i=1}^n)$ (see \eqref{eq:nuestcvpp}).
Let $\kappa\in(0,1)$. For any continuous proper prior $\prior$, the sequence of Bayes-assisted intervals $\mathcal T_{\kappa,\theta,n}=\calC_{\kappa,n}^\BA(\widehat \Delta_{\theta,n},\widehat\sigma^\Delta_{\theta,n}; \prior)$ forms a $(1-\kappa)$--\AsympCS with approximation rate $1/\sqrt{n\log{n}}$ for $\Delta_\theta$ and asymptotic Type-I error control. \looseness=-1
\label{prop:asympcsdeltappi}
\end{proposition}
Finally, for both PPI and PPI\texttt{++}, the confidence sequences $\mathcal R_{\delta,\theta,n}$ and $\mathcal T_{\alpha-\delta,\theta,n}$ are combined via a Minkowski sum to obtain a $(1-\alpha)$--\AsympCS for $g_\theta$ with approximation rate $1/\sqrt{n\log{n}}$ for $\mu$ and asymptotic Type-I error control, of the form

\begin{align}
\calC_{\alpha,\theta,n}^{g}=\left[ \widehat g_{\theta,n} \pm \left\{\frac{\widehat\sigma^\Delta_{\theta,n}}{\sqrt{n}} \sqrt{\log\left(\frac{n(2\pi\kappa^2)^{-1}}{\eta_n(\widehat \Delta_{\theta,n}/\widehat\sigma^\Delta_{\theta,n})^2}\right)}
+ \frac{\widehat\sigma^f_{\theta,n}}{\sqrt{N_n}} \sqrt{\frac{1+N_n\rho^2}{N_n\rho^2}\log\left(\frac{N_n\rho^2+1}{\delta^2}\right)} \right\}
  \right] \label{eq:avbappi_Cg}
\end{align}
where $\widehat g_{\theta,n}$ is either the PPI estimator \eqref{eq:ghatppi} or the PPI\texttt{++} estimator \eqref{eq:ghatppipp}.
Solving \cref{eq:ppconfidenceinterval} gives the confidence region for $\theta^\star$. In the case of the squared loss, $\calC_{\alpha,n}^{\avpp}$ is an interval, given by
\begin{align}
  \calC_{\alpha,n}^{\avpp}=\left[ \widehat\theta_n \pm \left\{\frac{\widehat\sigma^\Delta_{0,n}}{\sqrt{n}} \sqrt{\log\left(\frac{n}{2\pi\kappa^2\eta_n(\widehat \Delta_{0,n}/\widehat\sigma^\Delta_{0,n})^2}\right)}
  + \frac{\widehat\sigma^f_{0,n}}{\sqrt{N_n}} \sqrt{\frac{1+N_n\rho^2}{N_n\rho^2}\log\left(\frac{N_n\rho^2+1}{\delta^2}\right)} \right\}\right]
  \label{eq:avbappi_Ctheta_squared_loss}
\end{align}
where $\widehat\theta_n$ is either the PPI estimator \eqref{eq:thetahatppi} or the PPI\texttt{++} estimator \eqref{eq:thetahatppipp}.
%%%%%%% Section 6
% !TEX root = ppics_main.tex

\section{Experiments}
\label{sec:experiments}
We compare the PPI and PPI\texttt{++} \AsympCS procedures introduced in \cref{sec:abppi} -- with and without Bayes assistance -- to the \AsympCS relying solely on labelled data (obtained from \cref{thm:AsympCSlocalprioriid} and referred to as ``classical'') on several estimation problems.
Bayes-assisted methods are annotated with (G), (L), or (T) to indicate Gaussian, Laplace, or Student-t priors with mean zero and scale depending on the task and reported in the Supplementary Material.
For the Student-t prior, we set the degrees of freedom to $2$ in all experiments.
Since PPI is motivated by settings with scarce labelled data and abundant unlabelled data, we consider the following experimental setting: labelled data arrive sequentially, i.e., $n = 1,2,\ldots$, while a large unlabelled dataset is available from the start, i.e., $N_n = N$ for all $n$, with $N \gg n$ large enough to exclude any uncertainty on the measure of fit $m_\theta$.
As discussed by \citet{Cortinovis2025}, this simplifies the comparison between non-assisted and Bayes-assisted PPI, as it rules out any potential loss of efficiency due to the Minkowski sum \eqref{eq:avbappi_Cg}, thereby isolating the effect of the Bayes correction on the CS procedure.
For synthetic data, we set $N = \infty$ to guarantee the simplification holds.
For real data, we empirically verify that $N$ is large enough to justify this assumption by confirming that anytime validity is preserved -- specifically, that the cumulative miscoverage rate remains below the chosen threshold $\alpha = 0.1$ for all $n$.
As with CLT-based CIs, the $n$ at which one starts counting the cumulative miscoverage rate of an asymptotic CS is inherently arbitrary;
unless otherwise stated, we choose $n = 40$, as we empirically find this to be a reasonably small labelled sample size at which the KMT coupling generally provides a good approximation. \looseness=-1

\subsection{Synthetic data}\label{sec:synthetic_data}
The synthetic experiments below follow a general structure: we start with $N = \infty$ unlabelled samples $\{\widetilde{X}_j\}_{j=1}^N \iidsim \mathbb{P}_X$ and successively sample $n$ labelled observations $\{(X_i, Y_i)\}_{i=1}^n \iidsim \mathbb{P}$ with the goal of estimating the mean $\theta^\star = \mathbb{E}[Y]$.
We compare methods in terms of the average interval volume as $n$ increases across repetitions, and report the associated cumulative miscoverage rate in \cref{supp:synthetic_data}.

\begin{figure}
    \centering
    \includegraphics[width=\textwidth]{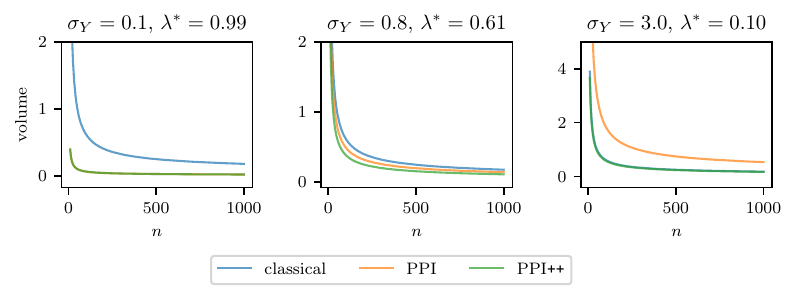}
    \caption{
        Noisy predictions study. The left, middle and right panels show average interval volume over $1000$ repetitions as a function of the labelled sample size $n$ for noise levels $\sigma_Y \in \{0.1, 0.8, 3.0\}$.
    }
    \label{fig:simulation_noisy}
\end{figure}

\paragraph{Noisy predictions.}
This experiment demonstrates that our method can adapt to varying correlation levels between predictions and true labels by using the PPI\texttt{++} estimator \eqref{eq:deltahatpowertuning}.
We sample $Y_i \iidsim \mathcal{N}(0, 1)$, so that $\theta^\star = \E[Y] = 0$.
The prediction rule is defined as $f(X_i) = Y_i + \epsilon_i$, where $X_i$ is only used for indexing and $\epsilon_i \iidsim \mathcal{N}(0, \sigma_Y^2)$, with the noise level $\sigma_Y \in \{0.1, 0.8, 3\}$.
In this case, the optimal control-variate parameter is given by $\lambda^\star_\theta = \lambda^\star = \cov(Y, f(X))/\var(f(X)) = (1 + \sigma_Y^2)^{-1}$, which decreases with $\sigma_Y$.
\Cref{fig:simulation_noisy} compares the interval volume achieved by classical and non-assisted CS procedures as a function of $n$, while results under informative priors are reported in \cref{supp:synthetic_data}.
For small noise levels, PPI and PPI\texttt{++} achieve similar performance, and greatly outperform classical inference.
As the noise level grows, the machine learning predictions become less informative and standard PPI loses ground to the classical CS.
By contrast, PPI\texttt{++} adapts to the noise level and always performs similarly to, or better than, the other baselines.

\begin{figure}[h]
    \centering
    \includegraphics[width=\textwidth]{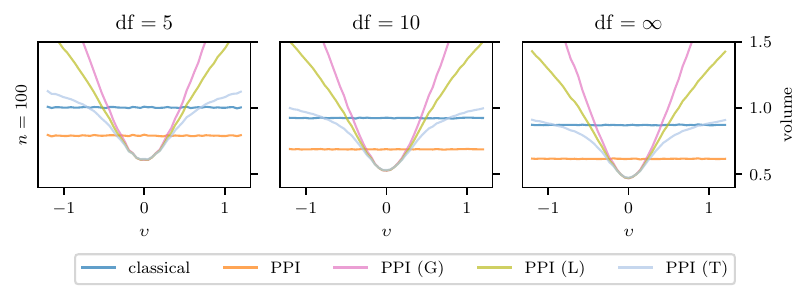}
    \caption{
        Biased predictions study. The left, middle and right panels show average interval volume over $100$ repetitions as a function of the bias level $\upsilon$ for $\mathrm{df} = 5, 10, \infty$.
    }
    \label{fig:simulation_biased}
\end{figure}

\paragraph{Biased predictions.}
This experiment illustrates the potential benefits of incorporating prior information into our method.
We sample $X_i \iidsim \mathcal{N}(0, 1)$ and $Y_i = X_i + \epsilon_i$, where $\epsilon_i \iidsim t_{\mathrm{df}}(0,1)$, so that $\theta^\star = \mathbb{E}[Y] = 0$.
The prediction rule is defined as $f(X_i) = X_i + \upsilon$, where $\upsilon \in \mathbb{R}$ controls its bias level.
For all $\upsilon$, $\lambda^\star =  1$, so that PPI and PPI\texttt{++} coincide.
We vary $\upsilon$ between $-1.2$ and $1.2$, and $\mathrm{df} \in \{5, 10, \infty\}$ to study the impact of bias level and noise distribution on the \AsympCS procedures.
\Cref{fig:simulation_biased} compares the average interval volumes at $n = 100$ as a function of $\upsilon$ for each value of $\mathrm{df}$.
Classical inference and non-assisted PPI volumes remain essentially constant across bias levels, reflecting their lack of prior information, and with the latter consistently outperforming the former by leveraging imputed predictions.
On the other hand, the volume of the Bayes-assisted procedures varies widely with the bias level $\upsilon$: the volume is reduced for small $\upsilon$, but grows with $|\upsilon|$ as the priors become increasingly misspecified.
Notably, the volume under the Gaussian prior inflates the fastest with $|\upsilon|$, while heavier-tailed Laplace and Student-t priors offer comparatively greater robustness.
These conclusions hold for all values of $\mathrm{df}$, which controls the accuracy of the KMT coupling approximation for a given $n$.
Coverage results in \cref{supp:synthetic_data} show that, while smaller values of $\mathrm{df}$ lead to slightly worse coverage, the approximation quality is overall satisfactory in this example. \looseness=-1

\subsection{Real data}\label{sec:real_data}
We evaluate our method on several real-world datasets, which are described in \cref{supp:datasets}.
While each dataset is, in principle, static (providing label/prediction pairs $\{(Y_i, f(X_i))\}_{i=1}^{N+n_1}$), we simulate an online setting akin to \cref{sec:synthetic_data} by randomly splitting the data into a labelled set of size $n_1$, which serves as a labelled data stream, and an unlabelled set of size $N$.

\begin{figure}
    \centering
    \includegraphics[width=\textwidth]{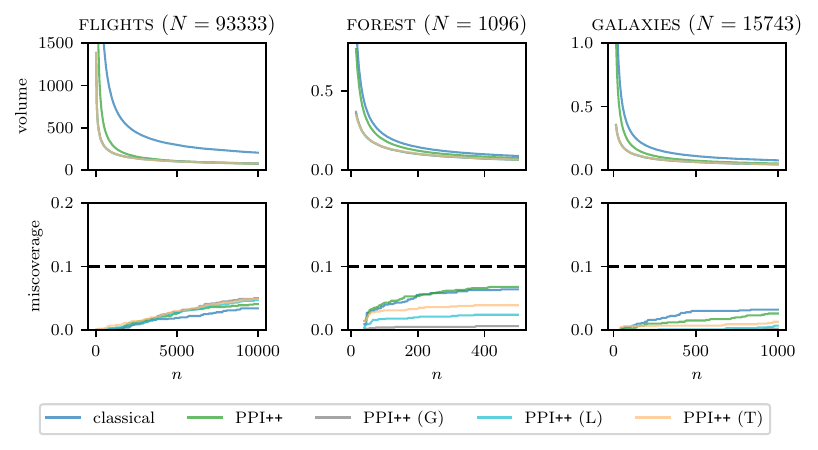}
    \caption{
        Mean estimation. The top and bottom rows show the average interval volume and cumulative miscoverage rate over $1000$ repetitions for the \textsc{flights}, \textsc{forest}, and \textsc{galaxies} datasets.
    }
    \label{fig:real_data}
\end{figure}

\Cref{fig:real_data} compares classical and PPI\texttt{++} \AsympCS procedures on the \textsc{flights}, \textsc{forest}, and \textsc{galaxies} datasets, where the goal is mean estimation.
By taking advantage of the unlabelled data, PPI methods consistently yield smaller regions than the classical counterpart, while maintaining reliable coverage.
Moreover, Bayes-assisted approaches further improve efficiency for moderate labelled sample sizes, as the quality of the predictions is generally high in these datasets.

\Cref{fig:other_estimation} reports results for three additional estimation tasks: linear regression (\textsc{census}), logistic regression (\textsc{healthcare}), and quantile estimation (\textsc{genes}).
For the first two tasks, the same conclusions as for mean estimation hold: PPI methods consistently outperform classical inference, with Bayes-assisted approaches providing an additional efficiency boost.
For the quantile estimation task, non-assisted PPI still improves over classical inference by leveraging the machine learning predictions;
however, the Bayes-assisted methods yield larger regions than the other approaches, reflecting lower prediction quality in this dataset.

%%%%%%% Section 7
% !TeX root = ppics_main.tex

\section{Discussion}
\label{sec:discussion}

We extended the PPI framework to the sequential setting via asymptotic confidence sequences, which further allow for seamless integration of prior information about the quality of the auxiliary predictions.
However, several directions merit further investigation. The results developed here are for scalar parameter values $\theta$.
Extensions to multivariate settings are discussed in \cref{supp:multivariate}, building on earlier work by~\citet[\S B.10]{WaudbySmith2024a}. In the non-assisted case, we focused on asymptotic confidence sequences of the form \eqref{eq:NonBayesianConfidenceSequence}, but other options are possible.
In particular, as discussed in \cref{supp:extendedville}, the parameter-free CS proposed by \citet{Wang2023}, which is based on an improper prior, may be used as an exact reference CS in place of \cref{eq:NonBayesianConfidenceSequence}.

The \AsympCS derived in this paper are asymptotically valid for \iid data under mild, nonparametric assumptions.
Promising directions include extensions to non-\iid observations, as well as the development of \emph{nonasymptotic}, nonparametric Bayes-assisted confidence sequences under stricter assumptions (e.g., bounded means), building on the work of \citet{WaudbySmith2024}. In the non-assisted case, the parameter $\rho$ was assumed to be fixed.
\citet[§2.5]{WaudbySmith2024a} considered delayed-start sequences $\calC_{\alpha,t}(m)$ that may depend on the start time $m$; this includes allowing the tuning parameter $\rho$ to depend on $m$.
Their asymptotic Type-I error control result, derived under assumptions similar to those used here, also applies in our setting. Another interesting direction would be to adapt similar ideas to the Bayes-assisted construction.

PPI \AsympCS procedures share the computational considerations of their fixed-time counterparts.
Beyond mean estimation (e.g., \cref{fig:other_estimation}), they typically require constructing a grid over $\theta$.
When the marginal density $\eta_t$ is not available in closed form (e.g., for the Student-$t$ prior), the Bayes-assisted version involves numerical integration.
If computation is a concern, the Laplace prior offers a good compromise: it has heavier tails than the Gaussian while still admitting a closed-form expression for $\eta_t$. \looseness=-1

\newpage

\begin{ack}
Valentin Kilian is supported by the Clarendon Funds Scholarship. Stefano Cortinovis is supported by the EPSRC Centre for Doctoral Training in Modern Statistics and Statistical Machine Learning (EP/S023151/1). The authors thank the reviewers for their time and valuable feedback, especially the suggestion to incorporate a discussion on Type-I error control. %They also would like to thank Daniel de Vassimon Manela for shining so brightly.
\end{ack}

\bibliographystyle{unsrtnat}

\clearpage

\newpage
\appendix

\renewcommand{\thesection}{S\arabic{section}}
\renewcommand{\thetheorem}{S\arabic{theorem}}
\renewcommand{\theproposition}{S\arabic{proposition}}
\renewcommand{\thelemma}{S\arabic{lemma}}
\renewcommand{\thedefinition}{S\arabic{definition}}
\renewcommand{\theassumption}{S\arabic{assumption}}
\renewcommand{\thefigure}{S\arabic{figure}}
\renewcommand{\thetable}{S\arabic{table}}
\renewcommand{\theequation}{S\arabic{equation}}

\begingroup % To keep \centering local
  
  \vskip 0.1in % From \@maketitle

  % --- Top Title Bar ---
  \hrule height 4pt
  \vskip 0.25in
  \vskip -\parskip

  % --- The Title ---
  \centering
  {\LARGE\bf Supplement to Anytime-valid, Bayes-assisted,\\Prediction-Powered Inference\par}

  % --- Bottom Title Bar ---
  \vskip 0.29in
  \vskip -\parskip
  \hrule height 1pt
  \vskip 0.09in

\endgroup
\medskip
  
  The supplementary material is organised as follows. \cref{supp:background} gives additional background on strong laws and couplings, and confidence sequences. \cref{supp:secondary_results} states secondary results and their proofs. \cref{supp:proof} presents the proofs of the main theorems and propositions. \cref{supp:multivariate} extends the results to the multivariate setting. \cref{supp:mean} gives specific expressions for the case of prediction-powered mean estimation. \cref{supp:experimental_details} details the experimental setup used in the main text. \cref{supp:addexp} presents further experiments. Finally, \cref{supp:extendedville} discusses a parameter-free, non-assisted, AsympCS, and provides some additional comparisons.

For clarity, all sections, theorems, propositions, and lemmas in the supplementary material are prefixed with ``S" to distinguish them from those in the main text.

%\tableofcontents
\startcontents[apx]

\section*{Contents}
\printcontents[apx]{0}{1}{}

\newpage

% !TEX root = ppics_supp.tex

 \section{Additional background}
 \label{supp:background}
 
 \subsection{Asymptotic theory of partial sums}

\subsubsection{Iterated logarithm and Marcinkiewicz-Zygmund strong laws}

\begin{theorem}(Iterated Logarithm Law~\cite[Theorem 8.5.2]{Durrett2019})
Let $(Y_t)_{t\geq 1}$ be \iid random variables with zero mean and unit variance.
Let $S_{t}=\sum_{i=1}^t Y_i$.
Then,
$$
    \underset{t\rightarrow\infty}{\lim\sup} \frac{\vert S_{t}\vert}{\sqrt{2t\log\log t}}=1 \text{~~~a.s.},
$$
which implies
$$
    \left\vert \frac{S_t}{t} \right\vert = O\left(  \sqrt{\frac{\log\log t}{t}}\right) \text{~~~a.s. as }t\to\infty.
$$
\end{theorem}

\begin{theorem}{(Marcinkiewicz-Zygmund strong law of large numbers~\cite[Theorem 2.5.12]{Durrett2019})}
\label{thm:MZSLLN}
Let $(Y_t)_{t\geq 1}$ be \iid random variables with zero mean and $\mathbb{E}\vert Y_1\vert^p<\infty$ for some $1<p<2$.
Let $S_t=\sum_{i=1}^tY_i$.
Then,
$$
    \frac{S_t}{t^{1/p}}\rightarrow 0 \text{~~~a.s. as }t\to\infty.
$$
\end{theorem}

\subsubsection{Strong approximations}

The following strong invariance result, attributed to Koml\'os, Major and Tusnady (KMT)~\cite{Komlos1975, Major1976} shows that the partial sums of i.i.d. random variables can be approximated almost surely by a Brownian motion path. The following theorem is from \citet[Theorem 3.2]{Csorgo1984}.

\begin{theorem}[KMT strong coupling~\cite{Komlos1975, Major1976}]
\label{thm:KMT2}
Let $(Y_t)_{t\geq 1}$ be \iid random variables with zero mean and unit variance such that $\mathbb{E}\vert Y_1\vert^q<\infty$ for some $q>2$. Then, there exists a Brownian motion B  such that, if we write $S_{t}=\sum_{i=1}^t Y_i$, we have
$$
    S_{t}-B_{t} =o(t^{1/q})  \text{~~~a.s.~as~} t\rightarrow \infty.
$$
\end{theorem}

KMT strong coupling has been extended by \citet{Einmahl1987} to random vectors (see also \cite[Section B11]{WaudbySmith2024b}).

\begin{theorem}(Multivariate KMT strong coupling \cite{Einmahl1987})
\label{thm:multiKMT}
Let $(Y_t)_{t\geq 1}$ be \iid random vectors in $\mathbb{R}^d$ with zero mean, covariance matrix $\Sigma$, and such that $\mathbb{E}\vert\vert Y_1\vert\vert^q<\infty$ for some $q>2$. Let $S_{t}=\sum_{i=1}^t Y_i$. Then, there exists a standard multivariate Brownian motion B  such that,
\[
    \Sigma^{-{1/2}}S_{t}-B_{t} =o(t^{1/q})  \text{~~~a.s.~as~} t\rightarrow \infty.
\]
\end{theorem}

\subsection{Confidence sequences}

\subsubsection{Confidence intervals vs.\ confidence sequences}

Let $(X_t)_{t\ge 1}$ be an observed data stream and let $\mu\in\bbR$ denote a fixed but unknown parameter (e.g., a mean). Write $\calF_t=\sigma(X_{1:t})$ for the natural filtration, and let $\alpha\in(0,1)$ be a pre-specified error probability (so the confidence level is $1-\alpha$).

\begin{figure}[h]
    \centering
    \includegraphics[width=0.6\textwidth]{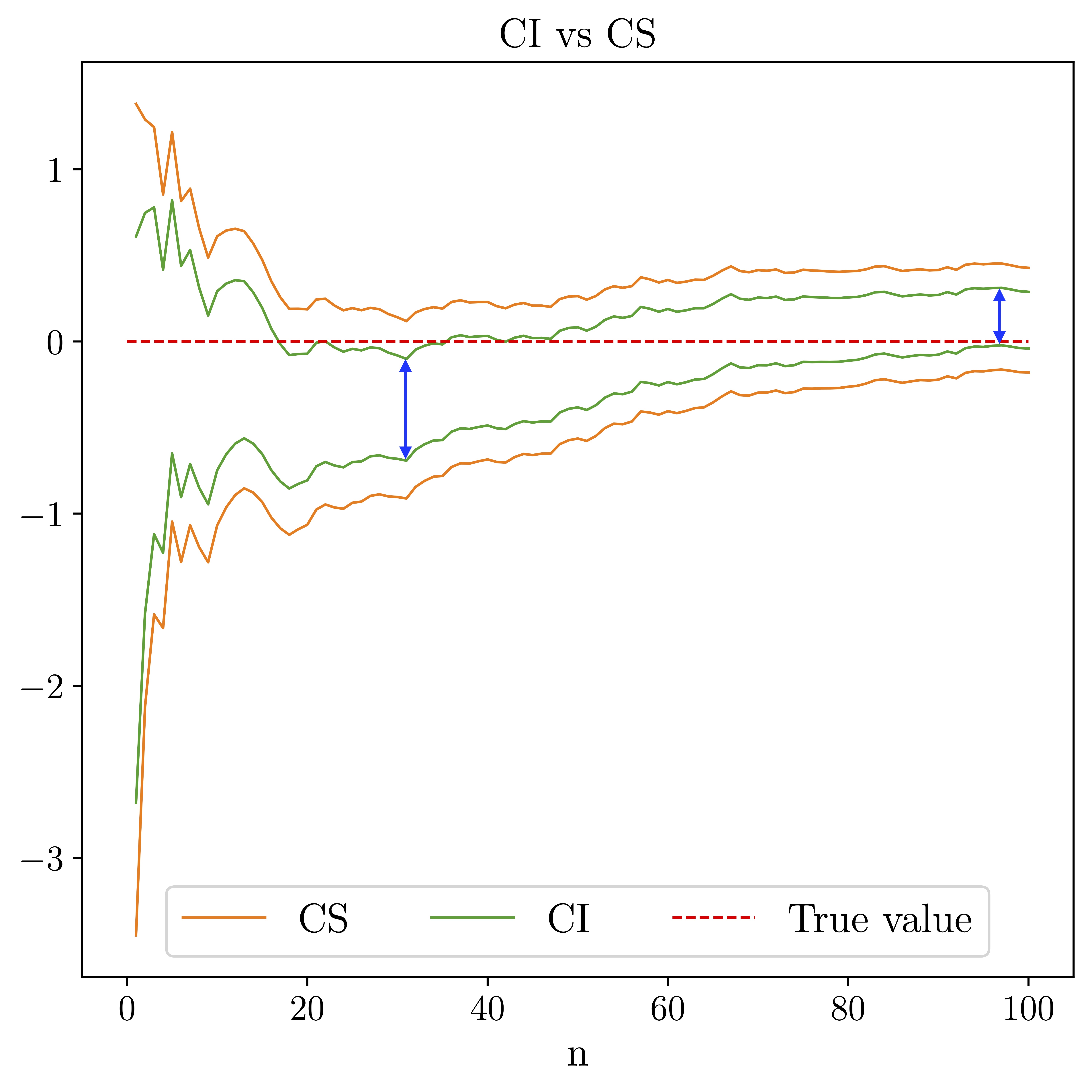}
    \caption{Comparison of fixed-time confidence intervals (CIs) and a confidence sequence (CS) for data from $\mathcal{N}(0,1)$. Two fixed-time CIs at different sample sizes happen to be disjoint (highlighted), illustrating that marginal coverage at each $t$ does not imply simultaneous coverage over $t$. The CS is more conservative at small $t$, but its coverage holds uniformly over all $t$.}
    \label{fig:CIvsCS}
\end{figure}

\paragraph{Fixed-time confidence intervals.}
A \emph{(fixed-time) confidence interval} (CI) for $\mu$ at time $t$ is an $\calF_t$-measurable random set $\calC_{\alpha,t}\subseteq\bbR$ such that
\[
\Pr\!\big(\mu\in\calC_{\alpha,t}\big)\ \ge\ 1-\alpha.
\]
This guarantee is \emph{marginal in $t$}: it holds for any chosen, deterministic $t$, but it need not be valid if $t$ is selected after looking at the data (e.g., by continual monitoring or a data-dependent stopping rule). In particular, for a general $\calF_t$-stopping time $\tau$,
\[
\Pr\!\big(\mu\in\calC_{\alpha,\tau}\big)\quad\text{can be }<\ 1-\alpha,
\]
unless the procedure is explicitly designed to be valid under optional stopping. Moreover, the family $(\calC_{\alpha,t})_{t\ge1}$ of fixed-time CIs need not be nested across $t$; disjoint intervals at different sample sizes can occur with positive probability (see \cref{fig:CIvsCS}), illustrating the lack of any simultaneous-in-time guarantee.

\paragraph{Confidence sequences.}
A \emph{confidence sequence} (CS) at level $1-\alpha$ is a sequence of $\calF_t$-measurable random sets $(\calC_{\alpha,t})_{t\ge1}$ such that
\[
\Pr\!\big(\mu\in \calC_{\alpha,t}\ \text{for all } t\ge1\big)\ \ge\ 1-\alpha.
\]
Equivalently,
\[
\Pr\!\Big(\sup_{t\ge1}\mathbf{1}\{\mu\notin \calC_{\alpha,t}\}=1\Big)\ \le\ \alpha.
\]
The quantifier ``for all $t$'' lies \emph{inside} the probability, yielding \emph{uniform-in-time} (a.k.a.\ anytime-valid) coverage. A key consequence is validity under arbitrary data-dependent stopping: for every (a.s.\ finite) stopping time $\tau$,
\[
\Pr\!\big(\mu\in \calC_{\alpha,\tau}\big)\ \ge\ 1-\alpha.
\]
Thus CSs support continual monitoring and sequential decision-making without inflating error rates. Practically, CSs are typically wider than fixed-time CIs at the same $t$ (especially early on) because they control the \emph{maximum} over all times; widths often shrink with $t$ and can approach classical rates up to iterated-logarithm factors.

Early examples of CSs go back to sequential analysis~\cite{Robbins1970a,Lai1976}, and modern constructions often proceed via nonnegative supermartingales/test martingales and time-uniform concentration inequalities.

\subsubsection{Nonnegative supermartingale and Ville's inequality}

\begin{definition}[Nonnegative supermartingale]
    $M=(M_{t})_{t\geq1}$ is a nonnegative supermartingale (NSM) with respect to the filtration $(\mathcal{F}_{t})_{t\geq1}$ if $M_{t}\geq0$ a.s, $\mathbb{E}[M_{t}]<\infty$ for all $t \geq 1$, and
    \begin{equation*}
        \mathbb{E}[M_{t+1} \mid \mathcal{F}_{t}] \leq M_{t}\text{ a.s.}
    \end{equation*}
    If there is equality, then $M$ is a nonnegative martingale.
\end{definition}

\begin{proposition}[Ville's inequality~\citep{Ville1939}]
    Let $(M_{t})_{t\geq1}$ be a nonnegative supermartingale.
    For any constant $c>0$,
    \begin{equation*}
        \Pr\left(  \sup_{t\geq1}M_{t}\geq c\right)  \leq\frac{\mathbb{E}[M_{1}]}{c}
    \end{equation*}
\end{proposition}
Ville's inequality can be seen as a generalisation of Markov's inequality. We have the following direct corollary.

\begin{proposition}
    \label{th:VilleCorollary}
    Let $(M_{t})_{t\geq1}$ be a nonnegative supermartingale.
    For any $\alpha\in(0,1)$,
    \begin{equation*}
        \Pr\left(  M_{t}\leq\frac{\mathbb{E}[M_{1}]}{\alpha}\text{ for all } t\geq1\right) \geq 1-\alpha.
    \end{equation*}
\end{proposition}

\subsubsection{Method of mixture}

Under the appropriate conditions, mixtures of martingales remain martingales:

\begin{proposition}[Lemma B1, \cite{Chugg2024}]
\label{thm:mixture}
Let $\{(M_t(\mu'))_{t\in\mathbb{N}}, \mu'\in\bbR\}$ be a family of (super)martingales on a filtered probability space $(\Omega, \mathcal{A}, (\mathcal{F}_t)_{t\in\mathbb{N}},\Pr)$, indexed by $\mu'$ in a measurable space $(\bbR, \mathcal{B})$, such that
\begin{enumerate}
\item each $M_t(\mu')$ is $\mathcal{F}_t \otimes \mathcal{B}$-measurable; and
\item each $\mathbb{E}\left[M_t(\mu') \mid \mathcal{F}_{t-1}\right]$ is $\mathcal{F}_{t-1} \otimes \mathcal{B}$-measurable.
\end{enumerate}
Let $\prior$ be a finite measure on $(\bbR, \mathcal{B})$ such that for all $n$,

$$
\Pr \otimes  \prior\text {-almost everywhere } M_t(\mu') \geq 0, \quad \text { or } \quad \mathbb{E}_{\mu' \sim \prior} \mathbb{E}\left[\left|M_t(\mu')\right|\right]<\infty
$$

Then the mixture $\left(\tilde M_t\right)_{t \in \mathbb{N}}$, where $\tilde M_t=\mathbb{E}_{\mu' \sim \prior} M_t(\mu')$, is also a (super)martingale.
\end{proposition}

This is useful as it leads to the \textit{method of mixtures} : if we have a family of nonnegative supermartingale (say) of the form  $M_t(\mu')$ for $\mu' \in \mathbb{R}$ which satisfy conditions 1 and 2 above and a mixture distribution $\prior$ satisfying the assumptions of \cref{thm:mixture}, then we can conclude that $\int_{\mu'\in\mathbb{R}}M_t(\mu')d\prior(\mu')$ is also a supermartingale, and thus Ville's inequality gives for any $\alpha\in(0,1)$

\begin{equation}
\label{eq:methodofmixture}
\Pr\left( \int_{\mu'\in\mathbb{R}}M_t(\mu')d\prior(\mu')  \leq \frac{1}{\alpha}\text{ for all } t\geq1\right) \geq 1-\alpha.
\end{equation}

The method of mixtures dates back at least to \citet{Ville1939} and was developed in the context of sequential analysis by \citet{Wald1945}. It was then systematised and popularised by Darling and Robbins in the late 1960s, by Robbins and Siegmund in a series of papers culminating in \cite{Robbins1970a}, and by \citet{Lai1976}. The method of mixtures has found many applications, including confidence sequences~\citep{Lai1976,Balsubramani2015, Kaufmann2021, Howard2021, WaudbySmith2024a}, PAC-Bayes analysis \citep{Haddouche2023, Chugg2024}, anytime-valid testing \citep{Ramdas2023}, and A/B testing \citep{Johari2017}, to name but a few. 
% !TEX root = ppics_supp.tex

\section{Secondary results}
\label{supp:secondary_results}

\subsection{Strong coupling with \iid Gaussian}

The following proposition follows from KMT strong approximation (see \cref{thm:KMT2}). It will be used in the proofs of \cref{thm:AsympCSglobalprioriid} and \cref{prop:strassencvestimator}.

\begin{proposition}
Let $\xi_1,\xi_2\ldots$ be \iid random variables with mean $\mu$ and variance $\sigma^2$ such that $\mathbb{E}\vert \xi_1\vert^q<\infty$ for some $q>2$.  Let $(N_n)_{n\geq 1}$ be a strictly increasing sequence of positive integers with $N_n \geq n$. Let $r\in(0,1]$ and assume $|\frac{n}{N_n}-r|=O(1/n^{1-a})$ with $0<a<2/q$. Then, there exists a sequence of \iid Gaussian random variables $(W_i)_{i\geq 1}$ with mean $\mu$ and variance $r\sigma^2$ such that
$$
\frac{1}{N_n}\sum_{i=1}^{N_n} \xi_i=\frac{1}{n}\sum_{i=1}^n W_i + o\left(\frac{1}{n^{1-1/q}}\right)\text{ a.s. as }n\to\infty.
$$
\label{prop:KMT3}
\end{proposition}

\begin{proof}
By \cref{thm:KMT2}, there exists a Brownian motion $B$ such that, a.s.~as $n\to\infty$,
\begin{align}
\sum_{i=1}^{N_n} \frac{\xi_i-\mu}{\sigma} &= B_{N_n} + o\left(N_n^{1/q}\right)\nonumber\\
&=  B_{N_n} + o\left(n^{1/q}\right)\nonumber\\
&= \frac{N_n}{n} r B_{n/r} + (B_{N_n}-\frac{N_n}{n} r B_{n/r})+o\left(n^{1/q}\right)\label{eq:KMTproof1}.
\end{align}
We have
$$
B_{N_n}-\frac{N_n}{n} r B_{n/r}=B_{N_n}-B_{n/r}+B_{n/r}(1-\frac{N_n}{n}r).
$$
$B_{N_n}-B_{n/r}$ is a zero-mean Gaussian random variable with variance
\begin{align*}
\var(B_{N_n}-B_{n/r})&=|N_n-n/r|\\
&=\frac{N_n}{r}\left|r-\frac{n}{N_n}\right|\\
&=O(n^a).
\end{align*}
By an upper tail inequality for Gaussian random variables, for any $\epsilon>0$,
$$
\Pr(|B_{N_n}-B_{n/r}|>\epsilon n^{1/q})\leq 2 \exp\left(-\frac{\epsilon^2 n^{2/q}}{\var(B_{N_n}-B_{n/r})}\right).
$$
For $n_0:=n_0(\epsilon)$ large enough, for all  $n>n_0$,
\begin{align*}
\exp\left(-\frac{\epsilon^2 n^{2/q}}{\var(B_{N_n}-B_{n/r})}\right)&\leq \exp\left(-\epsilon^2 n^{2/q-a}\right)\leq \frac{1}{n^2}.
\end{align*}
By comparison,
$$
\sum_{n\geq 1} \Pr\left(|B_{N_n}-B_{n/r}|>\epsilon n^{1/q}\right) <\infty.
$$
It follows from the Borel-Cantelli lemma that $|B_{N_n}-B_{n/r}|=o\left(n^{1/q}\right)$ a.s. as $n\to\infty$. Similarly, $B_{n/r}(1-\frac{N_n}{n}r)$ is a zero-mean Gaussian random variable with variance $\frac{n}{r}(1-\frac{N_n}{n}r)^2=O(n^{2a-1})=O(n^a)$. Using a similar proof, we obtain $B_{n/r}(1-\frac{N_n}{n}r)=o\left(n^{1/q}\right)$ a.s. as $n\to\infty$.
So, from \cref{eq:KMTproof1}, we obtain
$$
\frac{1}{N_n}\sum_{i=1}^{N_n}\xi_i=\frac{1}{n}(n\mu +\sigma r B_{n/r}) + o\left(\frac{1}{n^{1-1/q}}\right).
$$
We have,
\begin{align*}
n\mu+ \sigma r B_{n/r}=\sum_{i=1}^n \left[\mu+\sigma r(B_{i/r} -B_{(i-1)/r})\right]=\sum_{i=1}^n W_i,
\end{align*}
where $W_i = \mu + \sigma r(B_{i/r} -B_{(i-1)/r})$ are \iid Gaussian random variables with mean $\mu$ and variance  $r\sigma^2$. This completes the proof.
\end{proof}

The following lemma will be useful in the proof of \cref{prop:asympcvestimator}.

\begin{lemma}
\label{lem:lambda}
Let $(U_i,V_i)$, $i=1,\ldots,n$, be i.i.d. copies of a pair of random variables $(U,V)$. Assume $\mathbb{E}\vert U\vert^{2+\delta}$ and $\mathbb{E}\vert V\vert^{2+\delta}<\infty$ for some $0<\delta<1$. Let $\lambda^\star=\frac{\cov(U,V)}{\var(U)}$ and $\widehat\lambda=\frac{\widehat\cov((U_i,V_i)_{i=1}^n)}{\widehat\var((U_i)_{i=1}^n)}$.
%Take $\lambda^\star$ and $\widehat\lambda$ as defined in \cref{subsect:control} and assume additionally that $\mathbb{E}\vert U\vert^{2+\delta}$ and $\mathbb{E}\vert V\vert^{2+\delta}<\infty$ for some $1>\delta>0$, then 
Then $$\vert\lambda^\star-\widehat\lambda\vert=o(n^{-\frac{\delta}{2+\delta}})\text{ a.s. as }n\to\infty.$$
\end{lemma}

\begin{proof}
Using the mean value theorem, we obtain
\begin{align*}
\vert\lambda^\star-\widehat\lambda\vert&\leq \frac{1}{\var(U)}\left\vert\frac{1}{n}\sum_{i=1}^n\left(U_iV_i-\bbE(UV)\right)\right\vert + \frac{\vert\bbE(U)\vert}{\var(U)}\left\vert\overline V-\bbE V\right\vert +  \frac{\vert\overline V\vert}{\var(U)}\left\vert\overline U-\bbE U\right\vert\\
&+ \widehat\cov((U_i,V_i)_{i=1}^n) K_1 \left\vert\frac{1}{n}\sum_{i=1}^n(U_i^2-\bbE(U^2))\right\vert + \widehat\cov((U_i,V_i)_{i=1}^n) K_1K_2\left\vert\overline U-\bbE U\right\vert
\end{align*}
where $K_1$ and $K_2$ are two constants, independent of $n$. By assumption, $UV$ and $U^2$ have finite moments of order $1+\delta/2$  and $U$ and $V$ have finite moments of order $2+\delta$  (and so of order $1+\delta/2$) thus we can apply \cref{thm:MZSLLN} with $p=1+\delta/2$ to obtain the result. 
\end{proof}

\subsection{Confidence sequence for \iid Gaussian variables with known variance}
\label{sec:subsec:CSiidGaussian}

An important step in the proof of all our results is the derivation of an exact confidence sequence for \iid Gaussian variables with known variance. The non-assisted confidence sequence is a well-known result that can be found for instance in \cite{Robbins1970a} or in the proof of Theorem 2.2 in \cite{WaudbySmith2024a}:

\begin{theorem}
\label{thm:NAiidgaussian}
Let $W_{i}\iidsim\mathcal{N}(\mu,\sigma^{2})$. For any parameter $\rho>0$, the sequence of intervals defined as
\begin{equation}
\calC^\NA_{\alpha,t}(\overline W_t,\sigma;\rho):=\left[ \overline W_t \pm \frac{\sigma}{\sqrt{t}} \sqrt{\left(1+\frac{1}{t\rho^2}\right)\log\left(\frac{t\rho^2+1}{\alpha^2}\right)} \right ]
\label{eq:ExactNonBayesianConfidenceSequence}
\end{equation}
 is an exact $1-\alpha$ confidence sequence for $\mu_{0}$, that is
\[
\Pr\left(  \mu\in \calC^\NA_{\alpha,t}(\overline W_t,\sigma;\rho)\text{ for all
}t\geq1\right)  \geq1-\alpha.
\]
\end{theorem}

We also establish such a confidence sequence under a general prior on the mean. While \citet[Proposition C.1]{Wang2023} propose an exact confidence sequence under a Gaussian prior, we extend their result to any continuous and proper prior.
\begin{theorem}
\label{thm:iidgaussian}
Let $W_{i}\iidsim\mathcal{N}(\mu,\sigma^{2})$ and let $\prior$ be a continuous and proper on $\mu/\sigma$, then

\[
\calC^\BA_{\alpha,t}(\overline W_t,\sigma;\prior)=\left[  \overline{W}_{t}\pm\frac{\sigma
}{\sqrt{t}}\sqrt{\log\left(  \frac{t}{2\pi\alpha^{2}}\right)  -2\log\eta_{t}\left( \frac{\overline{W}_{t}}{\sigma}\right)  }\right],
\]

where $\eta_t$ is defined in \cref{eq:marginaldensity}, is an exact $1-\alpha$ confidence sequence for $\mu_{0}$, that is
\[
\Pr\left(  \mu\in \calC^\BA_{\alpha,t}(\overline W_t,\sigma;\prior)\text{ for all
}t\geq1\right)  \geq1-\alpha.
\]

Of particular interest, we may consider the Gaussian prior $\mathcal{N}%
(\mu_0,\tau^{2})$, which gives
\[
\calC^\BA_{\alpha,t}(\overline W_t,\sigma;\Normal(\cdot;\mu_0,\tau^2))=\left[  \overline{W}_{t}\pm\frac{\sigma
}{\sqrt{t}}\sqrt{\log\left(  \frac{t\tau^{2}+1}{\alpha^{2}}\right)
+\frac{\left(  \overline{W}_{t}/\sigma-\mu_0\right)  ^{2}}{\left(  \tau
^{2}+1/t\right)  }}\right]  .
\]

\end{theorem}

\begin{proof}
The main idea is to apply the methods of mixture with the prior $\prior$. For each $t\geq1$, let $p_{t,\mu}(w_{1:t})$ be the joint density of the $W_{1},\ldots,W_{t}$ with respect to $\lambda^{\otimes t}$ where $\lambda$ is the Lebesgue measure, for some unknown mean parameter $\mu\in\bbR$, where $w_{1:t}=(w_{1},\ldots,w_{t})\in\mathbb{R}^{t}$. To simplify notations, we drop the subscript $t$ and simply write $p_{\mu}(w_{1:t})$ to denote this joint density. We have
$$
p_{\mu}(w_{1},\ldots,w_{t})=C(\sigma,w_{1},\ldots,w_{t})\times \phi_t\left(\frac{\overline w_t - \mu}{\sigma}\right)
$$
where $\overline w_t=\frac{1}{t}\sum_{i=1}^t w_i$, $C(\sigma,w_{1},\ldots,w_{t})$ does not depend on $\mu$ and $\phi_t(w)$ denotes the pdf of a zero-mean Gaussian random variable with variance $1/t$. 

  For any $\mu\in\mathbb{R}, w_{1:t}\in\mathbb{R}^n$, let

\begin{align}
    \tilde{M}_t(\overline w_{t}, \mu)&=\int_{\mathbb{R}}\frac{p_{\mu'}(w_{1:t})}{p_{\mu}(w_{1:t})}\prior\left(\frac{\mu'}{\sigma}\right)\frac{d\mu'}{\sigma}\label{eq:likelihood_ratio}\\
    &=\int_{\mathbb{R}}\frac{\phi_t\left(\frac{\overline w_t - \mu'}{\sigma}\right)}{\phi_t\left(\frac{\overline w_t - \mu}{\sigma}\right)}\prior\left(\frac{\mu'}{\sigma}\right)\frac{d\mu'}{\sigma}\\
    &=\frac{\eta_t(\overline w_t/\sigma)}{\phi_t((\overline w_t-\mu)/\sigma)}.
\end{align}

We define $M_t(\mu')=\frac{p_{\mu'}(W_{1:t})}{p_{\mu}(W_{1:t})}=\exp\left(-\frac{1}{2\sigma^2}\left(\sum_{i=1}^t(W_i-\mu')^2-\sum_{i=1}^t(W_i-\mu)^2\right)\right).$ We consider $(\mathcal{F}_t)_{t\in\mathbb{N}}$, the filtration adapted to the sequence of random variable $(W_t)_{t\in\mathbb{N}}$. For every $\mu'\in\mathbb{R}$ we have

\begin{align*}
M_t(\mu') & =M_{t-1}(\mu')\times\exp\left(\frac{1}{2\sigma^2}(\mu^2-{\mu'}^2)\right)\exp\left(-\frac{W_t}{\sigma^2}(\mu-\mu')\right),
\end{align*}
and so
\begin{align*}
\mathbb{E}\left[M_t(\mu') \mid \mathcal{F}_{t-1}\right]&=M_{t-1}(\mu')\times\exp\left(\frac{1}{2\sigma^2}(\mu^2-{\mu'}^2)\right)\mathbb{E}\exp\left(-\frac{W_t}{\sigma^2}(\mu-\mu')\right)\\
&=M_{t-1}(\mu')\times\exp\left(\frac{1}{2\sigma^2}(\mu^2-{\mu'}^2)\right)\exp\left(-\mu\frac{\mu-\mu'}{\sigma^2}+\frac{1}{2\sigma^2}(\mu-{\mu'})^2\right)\\
&=M_{t-1}(\mu').
\end{align*}
Hence, $\{(M_t(\mu'))_{t\in\mathbb{N}}, \mu'\in\mathbb{R}\}$ is a family of martingale with respect to the adapted filtration  $(\mathcal{F}_t)_{t\in\mathbb{N}}$.

$M_t(\mu')$ is clearly continuous in $W_i$ for all $i\in\{1,...,t\}$ and for $\mu'\in\mathbb{R}$. Hence, it is $\mathcal{F}_t \otimes \mathcal{B}(\mathbb{R})$-measurable. Similarly, $\mathbb{E}\left[M_t(\mu') \mid \mathcal{F}_{t-1}\right]=M_{t-1}(\mu')$ is $\mathcal{F}_{t-1} \otimes \mathcal{B}(\mathbb{R})$-measurable.

Finally, we have $M_t(\mu')\geq0$ $\Pr \otimes \lambda$-almost everywhere, where $\lambda$ is the Lebesgue measure on $\mathbb{R}$ (or any other measure dominated by the Lebesgue measure on $\mathbb{R}$). Then, by \cref{thm:mixture}, $(\tilde{M}_t(\overline W_{t},\mu))_{t\geq1}$ is a nonnegative martingale with respect to the adapted filtration  $(\mathcal{F}_t)_{t\in\mathbb{N}}$. So by Ville's inequality, we have :
$$
\Pr\left( \tilde{M}_t(\overline W_{t},\mu) \leq \frac{1}{\alpha}\text{ for all } t\geq1\right) \geq 1-\alpha.
$$

It follows that the sequence
$$
\calC^\BA_{\alpha,t}(\overline W_t,\sigma;\prior)=\left\{\mu ~ | ~ \tilde{M}_t(\overline W_t,\mu)\leq\frac{1}{\alpha}\right\}
$$
is an exact $1-\alpha$ confidence sequence for $\mu$. Finally,
\begin{align*}
\tilde{M}_t(\overline W_{t},\mu) \leq \frac{1}{\alpha}&\Longleftrightarrow\exp\left(\frac{t}{2\sigma^2}(\mu-\overline{W}_{t})^2\right)\eta_{t}\left(\frac{\overline{W}_{t}}{\sigma}\right)\frac{\sqrt{2\pi}}{\sqrt{t}}\leq\frac{1}{\alpha}\\
&\Longleftrightarrow \frac{t}{2\sigma^2}(\mu-\overline{W}_{t})^2+\log\left({\eta}_{t}\left(\frac{\overline{W}_{t}}{\sigma}\right)\frac{\sqrt{2\pi}}{\sqrt{t}}\right)\leq -\log(\alpha)\\
&\Longleftrightarrow (\mu-\overline{W}_{t})^2\leq\frac{\sigma^2}{t}\left(-2\log\left({\eta}_{t}\left(\frac{\overline{W_t}}{\sigma}\right)\right)-\log\left(\frac{2\pi\alpha^2}{t}\right)\right).\\
\end{align*}

\end{proof}

\subsection{Optimal control-variate parameter for PPI\texttt{++}}

The next proposition follows from an application of \cref{prop:asympcvestimator} to the PPI\texttt{++} estimators, identifying the value of the optimal control variate parameter in this case.

\begin{proposition}[Asymptotics for PPI\texttt{++}]
    \label{prop:asympppi}
    Assume that, for any $\theta\in\bbR$, $\E|\loss_{\theta}'(X_1,Y_1)|^2<\infty$ and $\E|\loss_{\theta}'(X_1,f(X_1))|^2<\infty$. Let
    \begin{align}
        \lambda_\theta^\star=\frac{\cov(\loss_{\theta}'(X_1,Y_1),\loss_{\theta}'(X_1,f(X_1)))}{\var(\loss_{\theta}'(X_1,f(X_1)))}.
        \label{eq:lambdastar}
    \end{align}
    Then, for any $\theta$, almost surely as $n\to\infty$,
    \begin{align}
        \widehat g^\PPpp_{\theta,n}& = \left[\frac{1}{n}\sum_{i=1}^{n} \loss_{\theta}'(X_i,Y_i)\right]-\lambda^\star_{\theta}\left(\left[\frac{1}{n}\sum_{i=1}^n \loss_{\theta}'(X_i,f(X_i))\right]- \widehat m_\theta  \right )
        +o\left(\sqrt{\frac{\log\log n}{n}}\right),\\
        \widehat \Delta^\PPpp_{\theta,n}&=\frac{1}{n}\sum_{i=1}^{n} \left(\loss_{\theta}'(X_i,Y_i)-\loss_{\theta}'(X_i,f(X_i))\right)  - (\lambda^\star_{\theta} -1 )\left( \frac{1}{n}\left [\sum_{i=1}^{n} \loss_{\theta}'(X_i,f(X_i))\right ]  - \widehat m_{\theta}\right)\nonumber\\
        &~~~~+o\left(\sqrt{\frac{\log\log n}{n}}\right).
    \end{align}
    Additionally, in the case of the squared loss,
    \begin{align}
        %\widehat\theta_n^\PP &= \frac{1}{n}\sum_{i=1}^{n}Y_i - \left( \frac{1}{n}\sum_{i=1}^{n} f(X_i) - \E[f(X)]     \right)+O\left(\sqrt{\frac{\log\log n}{n}}\right),\\
        \widehat\theta_n^\PPpp & = \frac{1}{n}\sum_{i=1}^{n}Y_i - \lambda^\star_{0}\left( \frac{1}{n}\sum_{i=1}^{n} f(X_i) - \frac{1}{N_n}\sum_{j=1}^{N_n} f(\tX_j)     \right)+o\left(\sqrt{\frac{\log\log n}{n}}\right),
    \end{align}
    with $\lambda^\star_{0}=\cov(Y_1,f(X_1))/\var(f(X_1))$.
\end{proposition}

% !TEX root = ppics_supp.tex

\section{Proofs}
\label{supp:proof}

\subsection{Proofs of  \cref{thm:AsympCSlocalprioriid} and \cref{thm:AsympCSlocalpriornoniid}}

\Cref{thm:AsympCSlocalprioriid} is a corollary of \cref{thm:AsympCSlocalpriornoniid}, so we start by proving \cref{thm:AsympCSlocalpriornoniid}.

\subsubsection{Proof of \cref{thm:AsympCSlocalpriornoniid}}

By assumption, we have, almost surely,
\[
\widehat\mu_t =\overline{W}_{t}+\varepsilon_{t},
\]
where $\varepsilon_{t}=o\left(\frac{1}{\sqrt{t\log t}}\right)$ and $\overline W_t=\frac{1}{t}\sum_{i=1}^t W_i$.

Using \cref{thm:NAiidgaussian}, the sequence of intervals $\calC^\NA_{\alpha,t}(\overline W_t,\sigma;\rho)=\calC^\NA_{\alpha,t}(\widehat\mu_t-\varepsilon_t,\sigma;\rho)=[\widehat\mu_t-L_{t}^{\ast},\widehat\mu_{t}+U_{t}^{\ast}]$, where
\begin{align*}
U_{t}^{\ast} &  =\frac{\sigma}{\sqrt{t}} \sqrt{\left(1+\frac{1}{t\rho^2}\right)\log\left(\frac{t\rho^2+1}{\alpha^2}\right)}-\varepsilon_{t}, \text{~and}\\
L_{t}^{\ast} &  =\frac{\sigma}{\sqrt{t}} \sqrt{\left(1+\frac{1}{t\rho^2}\right)\log\left(\frac{t\rho^2+1}{\alpha^2}\right)}+\varepsilon_{t},%
\end{align*}
\bigskip
is an exact confidence sequence for $\mu$. We have $\calC^\NA_{\alpha,t}(\widehat\mu_t,\widehat\sigma_t;\rho)=[\widehat\mu_t-L_t,\widehat\mu_t+U_t]$, where
\[
U_{t}=L_{t}=\frac{\widehat\sigma_t}{\sqrt{t}} \sqrt{\left(1+\frac{1}{t\rho^2}\right)\log\left(\frac{t\rho^2+1}{\alpha^2}\right)}.
\]

 Let $a_t=1/\sqrt{t \log t}$. Then,
\begin{multline*}
\frac{1}{a_t}\left(L_t-L_t^\star\right)=\frac{1}{a_t}\left[\frac{\widehat{\sigma}_t}{\sqrt{t}} \sqrt{\left(1+\frac{1}{t\rho^2}\right)\log\left(\frac{t\rho^2+1}{\alpha^2}\right)}\right.\\-\left.\frac{\sigma}{\sqrt{t}} \sqrt{\left(1+\frac{1}{t\rho^2}\right)\log\left(\frac{t\rho^2+1}{\alpha^2}\right)}\right]+o(1).
\end{multline*}
We have
$$
\begin{aligned}
& \frac{\widehat{\sigma}_t}{\sqrt{t}} \sqrt{\left(1+\frac{1}{t\rho^2}\right)\log\left(\frac{t\rho^2+1}{\alpha^2}\right)}-\frac{\sigma}{\sqrt{t}} \sqrt{\left(1+\frac{1}{t\rho^2}\right)\log\left(\frac{t\rho^2+1}{\alpha^2}\right)}\\
& \sim (\widehat\sigma_t-\sigma)\times\sqrt{\frac{\log t}{t}} = o\left(\frac{1}{\sqrt{t\log t}}\right).
\end{aligned}
$$
Hence $\frac{1}{a_t}\left(L_t-L_t^\star\right)=o(1)$. Similarly, $\frac{1}{a_t}\left(U_t-U_t^\star\right)=o(1)$. It follows that $\left(\calC_{\alpha, t}^{\NA}\right)$ is a $(1-\alpha)$--\AsympCS with approximation rate $1/\sqrt{t\log{t}}$.

 \subsubsection{Proof of  \cref{thm:AsympCSlocalprioriid}}

  In order to apply \cref{thm:AsympCSlocalpriornoniid} in this setting, we need \cref{condition1} to be satisfied for the \iid sequence $(Y_t)_{t\geq 1}$.
    By KMT strong coupling (\cref{thm:KMT2}), there exists a sequence of \iid Gaussian random variables $(W_i)_{i\geq 1}$ with mean $\mu$ and variance $\sigma^2$ such that, a.s.,
    $$
        \overline{Y}_t=\frac{1}{t} \sum_{i=1}^t W_i+\varepsilon_t
        \text{~~~where~~~}
        \varepsilon_t=o\left(\frac{1}{t^{1-1/(2+\delta)}}\right) =o\left(\frac{1}{\sqrt{t\log t}}\right).
    $$

  We also need to satisfied the condition on the variance. Under $\E[|Y|^{2+\delta}]<\infty$, the Marcinkiewicz-Zygmund strong law of large numbers (\cref{thm:MZSLLN}) with $p=1+\delta/2\in(1,2)$ yields a polynomial a.s. rate for $\overline Y_t$ and $\overline{Y^2_t}$; consequently $\vert\widehat\sigma_t-\sigma\vert=o(t^{-\gamma})$ for some $\gamma >0$, which implies
$$
\vert\widehat\sigma_t-\sigma\vert=o\left(\frac{1}{\log t}\right)\text{ a.s. as }t\to\infty.
$$

    The result follows.

\subsection{Proofs of \cref{thm:AsympCSglobalprioriid} and \cref{thm:AsympCSglobalpriornoniid}}

The following lemma is a direct consequence of Theorem 8.14(b) p. 242 \cite{Folland2013}.
\begin{lemma}
\label{thm:molifier}
Let $\prior$ be a proper and continuous probability density function on $\mathbb{R}^d$. Let $\left(  Z_{t}\right)  _{t\geq1}$ be a sequence of random vectors in $\bbR^d$, with
$Z_{t}\rightarrow c$ a.s. as $t\rightarrow\infty$. Let
$$
\eta_t(z)=\int_{\bbR^d} \Normal(z;\zeta,I_d/t)\prior(\zeta)d\zeta.
$$
Then
\[
{\eta}_{t}(Z_{t})\rightarrow\prior(c)\text{ almost surely as
}t\rightarrow\infty.
\]
\end{lemma}

\Cref{thm:AsympCSglobalprioriid} is a corollary of \cref{thm:AsympCSglobalpriornoniid}, so we start by proving \cref{thm:AsympCSglobalpriornoniid}.

\subsubsection{Proof of \cref{thm:AsympCSglobalpriornoniid}}

By assumption, we have, almost surely,
\[
\widehat\mu_t =\overline{W}_{t}+\varepsilon_{t},
\]
where $\varepsilon_{t}=o\left(\frac{1}{\sqrt{t\log t}}\right)$ and $\overline W_t=\frac{1}{t}\sum_{i=1}^t W_i$.

Using \cref{thm:iidgaussian}, the sequence of intervals $\calC^\BA_{\alpha,t}(\overline W_t,\sigma_;\prior)=\calC^\BA_{\alpha,t}(\widehat\mu_t-\varepsilon_t,\sigma;\prior)=[\widehat\mu_t-L_{t}^{\ast},\widehat\mu_{t}+U_{t}^{\ast}]$, where
\begin{align*}
U_{t}^{\ast} &  =\frac{\sigma}{\sqrt{t}}\sqrt{\log\left(  \frac{t}{2\pi
\alpha^{2}}\right)  -2\log{\eta}_{t}\left(  \frac{\widehat\mu_{t}-\varepsilon_{t}}{\sigma}\right)  }-\varepsilon_{t}, \text{~and}\\
L_{t}^{\ast} &  =\frac{\sigma}{\sqrt{t}}\sqrt{\log\left(  \frac{t}{2\pi
\alpha^{2}}\right)  -2\log{\eta}_{t}\left(  \frac{\widehat\mu_{t}-\varepsilon_{t}}{\sigma}\right)  }+\varepsilon_{t},%
\end{align*}
\bigskip
is an exact confidence sequence for $\mu$. We have $\calC^\BA_{\alpha,t}(\widehat\mu_t,\widehat\sigma_t;\prior)=[\widehat\mu_t-L_t,\widehat\mu_t+U_t]$, where
\[
U_{t}=L_{t}=\frac{\widehat\sigma_{t}}{\sqrt{t}}\sqrt{\log\left(  \frac
{t}{2\pi\alpha^{2}}\right)  -2\log{\eta}_{t}\left(  \frac{\widehat\mu_t }{\widehat\sigma_{t}}\right)}.
\]

 Let $a_t=1/\sqrt{t \log t}$. Then,
\begin{multline*}
\frac{1}{a_t}\left(L_t-L_t^\star\right)=\frac{1}{a_t}\left[\frac{\widehat{\sigma}_t}{\sqrt{t}} \sqrt{\log \left(\frac{t}{2 \pi \alpha^2}\right)-2 \log \eta_t\left(\frac{\widehat{\mu}_t}{\widehat{\sigma}_t}\right)}\right.\\-\left.\frac{\sigma}{\sqrt{t}} \sqrt{\log \left(\frac{t}{2 \pi \alpha^2}\right)-2 \log \eta_t\left(\frac{\widehat\mu_{t}-\varepsilon_t}{\sigma}\right)}\right]+o(1)
\end{multline*}
We have
$$
\begin{aligned}
& \frac{\widehat{\sigma}_t}{\sqrt{t}} \sqrt{\log \left(\frac{t}{2 \pi \alpha^2}\right)-2 \log \eta_t\left(\frac{\widehat{\mu}_t}{\widehat{\sigma}_t}\right)}-\frac{\sigma}{\sqrt{t}} \sqrt{\log \left(\frac{t}{2 \pi \alpha^2}\right)-2 \log \eta_t\left(\frac{\widehat\mu_{t}-\varepsilon_t}{\sigma}\right)} \\
& =\frac{\frac{\widehat{\sigma}_t^2}{t}\left[\log \left(\frac{t}{2 \pi \alpha^2}\right)-2 \log \eta_t\left(\frac{\widehat{\mu}_t}{\widehat{\sigma}_t}\right)\right]-\frac{\sigma^2}{t}\left[\log \left(\frac{t}{2 \pi \alpha^2}\right)-2 \log \eta_t\left(\frac{\widehat\mu_{t}-\varepsilon_t}{\sigma}\right)\right]}{\frac{\widehat{\sigma}_t}{\sqrt{t}} \sqrt{\log \left(\frac{t}{2 \pi \alpha^2}\right)-2 \log \eta_t\left(\frac{\widehat{\mu}_t}{\widehat{\sigma}_t}\right)}+\frac{\sigma}{\sqrt{t}} \sqrt{\log \left(\frac{t}{2 \pi \alpha^2}\right)-2 \log \eta_t\left(\frac{\widehat\mu_{t}-\varepsilon_t}{\sigma}\right)}} \\
& \sim (\widehat\sigma_t-\sigma)\times\sqrt{\frac{\log t}{t}} = o\left(\frac{1}{\sqrt{t\log t}}\right)
\end{aligned}
$$
as, by \cref{thm:molifier}, we have
$$
    {\eta}_{t}\left(   \frac{\widehat\mu_t }{\widehat\sigma_{t}}\right)  \rightarrow\prior\left(
    \frac{\mu}{\sigma}\right)  \text{~and~}  {\eta}_{t}\left(   \frac{\widehat\mu_{t}-\varepsilon_t }{\sigma}\right)  \rightarrow\prior\left(
    \frac{\mu}{\sigma}\right)   \text{~a.s. as~} t\rightarrow\infty.
$$
It follows that $\left(\calC_{\alpha, t}^{\BA}\right)$ is a $(1-\alpha)$--\AsympCS with approximation rate $1/\sqrt{t\log{t}}$.

\subsubsection{Proof of \cref{thm:AsympCSglobalprioriid}}

The result follows from \cref{thm:AsympCSglobalpriornoniid} by applying the same reasoning as for the proof of \cref{thm:AsympCSlocalprioriid}.

\subsection{Proof of \cref{thm:typeIerror}}

The idea of the proof is as follows. Recall that the intervals $(\calC_{\alpha,t})$ of interest are approximations of an exact CS $(\calC^\star_{\alpha,t})$. For any $\alpha'>\alpha$, the narrower exact CS $(\calC^\star_{\alpha',t})$ is eventually (a.s.) contained in $(\calC_{\alpha,t})$ for all large $t$. A standard sandwiching argument using this eventual containment yields the desired asymptotic Type-I control.\medskip

Let $a_t=1/\sqrt{t\log t}$. For each construction (non-assisted/Bayes-assisted), the sequence of intervals of interest are of the form  $\calC_{\alpha,t}=[\widehat\mu_t\pm U_{\alpha,t}]$ where
\[
U_{\alpha,t}=\frac{\widehat\sigma_t}{\sqrt{t}} \sqrt{\left(1+\frac{1}{t\rho^2}\right)\log\left(\frac{t\rho^2+1}{\alpha^2}\right)}
\]
for the non-assisted case, and
\[
U_{\alpha,t}=\frac{\widehat\sigma_{t}}{\sqrt{t}}\sqrt{\log\left(  \frac
{t}{2\pi\alpha^{2}}\right)  -2\log{\eta}_{t}\left(  \frac{\widehat\mu_t }{\widehat\sigma_{t}}\right)}
\]
for the Bayes-assisted case. $\calC_{\alpha,t}$ approximates a reference exact CS of the form $\calC^\star_{\alpha,t}=[\widehat\mu_t-L^\star_{\alpha,t},\widehat\mu_t+ U^\star_{\alpha,t}]$, where
\begin{align*}
U_{\alpha,t}^{\ast} &  =\frac{\sigma}{\sqrt{t}} \sqrt{\left(1+\frac{1}{t\rho^2}\right)\log\left(\frac{t\rho^2+1}{\alpha^2}\right)}-\varepsilon_{t}, \text{~and}\\
L_{\alpha,t}^{\ast} &  =\frac{\sigma}{\sqrt{t}} \sqrt{\left(1+\frac{1}{t\rho^2}\right)\log\left(\frac{t\rho^2+1}{\alpha^2}\right)}+\varepsilon_{t},%
\end{align*}
for the non-assisted case, and
\begin{align*}
U_{\alpha,t}^{\ast} &  =\frac{\sigma}{\sqrt{t}}\sqrt{\log\left(  \frac{t}{2\pi
\alpha^{2}}\right)  -2\log{\eta}_{t}\left(  \frac{\widehat\mu_{t}-\varepsilon_{t}}{\sigma}\right)  }-\varepsilon_{t}, \text{~and}\\
L_{\alpha,t}^{\ast} &  =\frac{\sigma}{\sqrt{t}}\sqrt{\log\left(  \frac{t}{2\pi
\alpha^{2}}\right)  -2\log{\eta}_{t}\left(  \frac{\widehat\mu_{t}-\varepsilon_{t}}{\sigma}\right)  }+\varepsilon_{t},%
\end{align*}
for the Bayes-assisted case, where $\varepsilon_t=\widehat\mu_t-\overline W_t=o(a_t)$ a.s. does not depend on $\alpha$. In both cases, $(\calC^\star_{\alpha,t})_{t\geq 1}$ is an exact CS (\cref{thm:NAiidgaussian,thm:iidgaussian}); hence, for $E_\alpha^\star=\{\mu \in \calC^{\star}_{\alpha,t}\text{ for all }t\geq 1\}$,
$$
\Pr(E_\alpha^\star)\geq 1-\alpha.
$$
Additionally, $$U_{\alpha,t}\sim U^\star_{\alpha,t}\sim L^\star_{\alpha,t}\sim \sigma\sqrt{\frac{\log t}{t}}\text{ a.s. as }t\to\infty,$$
and, as shown in the proofs of \cref{thm:AsympCSlocalpriornoniid,thm:AsympCSglobalpriornoniid}, a.s.,
\begin{align}
U_{\alpha,t}-U^\star_{\alpha,t}=o(a_t)\text{ and }U_{\alpha,t}-L^\star_{\alpha,t}=o(a_t).
\label{eq:asympUalphavsUalphastar}
\end{align}

Let $\alpha'\in(\alpha,1)$. We now aim to show that, for some random, finite time $T_{\alpha'}$, $\calC^\star_{\alpha',t} \subseteq \calC_{\alpha,t}$ for all $t\geq  T_{\alpha'}$. We have
\begin{align}
U^\star_{\alpha,t}-U^\star_{\alpha',t}=\frac{(U^\star_{\alpha,t})^2-(U^\star_{\alpha',t})^2}{U^\star_{\alpha,t}+U^\star_{\alpha',t}}\sim\frac{\sigma^2/t \log \frac{(\alpha')^2}{\alpha^2}}{2\sigma\sqrt{\log t /t}}\sim c(\alpha,\alpha')a_t
\label{eq:asymptUalphavsUalphaprime}
\end{align}
a.s. as $t\to\infty$, where $c(\alpha,\alpha')=\sigma\log(\alpha'/\alpha)>0$. Similarly,
\begin{align}
L^\star_{\alpha,t}-L^\star_{\alpha',t}\sim c(\alpha,\alpha')a_t\text{ a.s. as }t\to\infty.
\label{eq:asymptLalphavsLalphaprime}
\end{align}
Combining \cref{eq:asymptLalphavsLalphaprime,eq:asymptUalphavsUalphaprime} with \cref{eq:asympUalphavsUalphastar}, we obtain, a.s.
\begin{align}
U_{\alpha,t}-U^\star_{\alpha',t}\sim c(\alpha,\alpha')a_t\\
L_{\alpha,t}-L^\star_{\alpha',t}\sim c(\alpha,\alpha')a_t.
\end{align}
So, there exists a collection of events $\Omega_0$ with $\Pr(\Omega_0)=1$ such that for every $\omega\in\Omega_0$, there is a finite $T_{\alpha'}(\omega)$ with
\begin{align}
U_{\alpha,t}(\omega)-U^\star_{\alpha',t}(\omega)\geq \frac{1}{2}c(\alpha,\alpha')a_t\\
L_{\alpha,t}(\omega)-L^\star_{\alpha',t}(\omega)\geq \frac{1}{2} c(\alpha,\alpha')a_t.
\end{align}
for all $t\geq T_{\alpha'}(\omega)$. Therefore, $\calC^\star_{\alpha',t} \subseteq \calC_{\alpha,t}$ for all $t\geq  T_{\alpha'}$. It follows, that for every $m\geq 1$,
$$
\Pr(\mu\in \calC_{\alpha,t} \text{ for all }t\geq m)\geq \Pr(E^\star_{\alpha'}\cap \{T_{\alpha'}\leq m\})\longrightarrow_{m\to\infty}\Pr(E^\star_{\alpha'})\geq 1-\alpha'.
$$
Hence, for any $\alpha'\in(\alpha,1)$, $\lim\inf_{m\to\infty} \Pr(\mu\in \calC_{\alpha,t} \text{ for all }t\geq m)\geq 1-\alpha'$ thus
$$
{\lim\inf}_{m\to\infty} \Pr(\mu\in \calC_{\alpha,t} \text{ for all }t\geq m)\geq 1-\alpha.
$$

\subsection{Proof of \cref{prop:asympcvestimator}}

Let
\begin{align*}
\epsilon_n&=\widehat \gamma^{\cvpp}-\widehat \gamma_{\lambda^\star}^{\cv}\\
&=\widehat \gamma^{\cvpp}-(\overline V - \lambda^\star (\overline U - \widehat\mu))\\
&=(\lambda^\star-\widehat\lambda)(\overline U-\mu)-(\lambda^\star-\widehat\lambda)(\widehat \mu -\mu).
\end{align*}
By the triangle inequality,
\begin{align*}
|\epsilon_n|\leq |\lambda^\star-\widehat\lambda|(|\overline U-\mu|+|\widehat \mu -\mu|).
\end{align*}
By the \cref{lem:lambda} we have 
$$
\vert\widehat\lambda-\lambda^\star\vert=o\left(n^{-2/(2+\delta)}\right)\text{ a.s. as }n\to\infty
$$
and, by the law of the iterated logarithm
\begin{align}
|\overline U-\mu|=O\left(\sqrt{\frac{\log\log n}{n}}\right),~~~|\widehat \mu-\mu|=O\left(\sqrt{\frac{\log\log n}{n}}\right)~~\text{ a.s. as }n\to\infty.
\end{align}

It follows that
\begin{align}
|\epsilon_n|=O\left(\frac{1}{n^{2/(2+\delta)}}\sqrt{\frac{\log\log n}{n}}\right)=o\left(\frac{1}{\sqrt{n\log n}}\right)~~\text{ a.s. as }n\to\infty.
\end{align}

\subsection{Proof of \cref{prop:strassencvestimator}}

We have
\begin{align}
\widehat \gamma_\lambda^{\cv}=\overline V - \lambda (\overline U - \mu)+\lambda(\widehat\mu-\mu).
\end{align}
The random variables $\overline V - \lambda (\overline U - \mu)$ and $\lambda(\widehat\mu-\mu)$ are independent and are both sample average of \iid random variables with finite moment of order $q$ for some $q=2+\delta>2$. Note that $1-1/q>1/2$. By KMT strong coupling (\cref{thm:KMT2}), there exist \iid Gaussian random variables $(G^{(1)}_i)_{i\geq 1}$ with mean $\gamma$ and variance $\var(V-\lambda U)$ such that
$$
\overline V - \lambda (\overline U - \mu)=\frac{1}{n}\sum_{i=1}^n G^{(1)}_i+ o\left(\frac{1}{n^{1-1/q}}\right)\text{ a.s. as }n\to\infty.
$$
If $r=0$, then $n/N_n=O(1/n^{1-a})$. By the law of the iterated logarithm,
\begin{align*}
|\lambda(\widehat\mu-\mu)|&=O\left(\sqrt{\frac{\log\log N_n}{N_n}}\right)=o\left(\frac{1}{\sqrt{n \log n}}\right)\text{ a.s. as }n \to\infty.
\end{align*}
If $r>0$, then, by \cref{prop:KMT3}, there exist \iid Gaussian random variables $(G^{(2)}_i)_{i\geq 1}$, independent of $(G^{(1)}_i)_{i\geq 1}$, with mean $0$ and variance $r\lambda^2 \var(U)$ such that,
$$
\lambda(\widehat\mu-\mu)=\frac{1}{n}\sum_{i=1}^n G^{(2)}_i+ o\left(\frac{1}{n^{1-1/q}}\right)\text{ a.s. as }n\to\infty.
$$
Setting $W^{\cv}_i=G^{(1)}_i$ if $r=0$ and $W^{\cv}_i=G^{(1)}_i+G^{(2)}_i$ if $r>0$ gives the Gaussian coupling \eqref{eq:CVEgammacoupling} for $\nu_\lambda^{\cv}$, as $\frac{1}{n^{1-1/q}}=o\left(\frac{1}{\sqrt{n\log n}}\right)$. From this, using \cref{eq:strongcvvscvpp}, we deduce the coupling \eqref{eq:CVEgammacoupling2} for $\widehat \gamma^{\cvpp}$, noting that
\begin{align}
\nu^{\cvpp} :=\nu_{\lambda^\star}^{\cv} =\var(V)\left[1-(1-r)\rho_{U,V}^2\right].
\end{align}
We have
$$
\frac{1}{n-2}\sum_{i=1}^n (V_i - \overline V-  \lambda (U_i-\overline U))^2\to \var(V-\lambda U)\text{ a.s.}
$$
and
$$
\frac{n\lambda^2}{N_n(N_n-1)}\sum_{j=1}^{N_n} (\widetilde U_j-\widehat\mu)^2\to r\lambda^2 \var(U)\text{ a.s.}
$$
therefore  $\widehat \nu_\lambda^\cv \to\nu_\lambda^\cv$ a.s. 
Finally, we show that $\widehat \nu^\cvpp$ is a consistent estimator of $\nu^\cvpp$. Set $\delta_i=V_i - \overline V-  \widehat\lambda (U_i-\overline U)$. We have $\delta_i= V_i-\widehat\alpha-\widehat\beta U_i$ where $\widehat\alpha=\overline V-\widehat\lambda\overline U$ and $\widehat\beta=\widehat\lambda$ are the least squares estimates, minimising $\sum_{i=1}^n (V_i - \alpha - \beta U_i)^2$. It is well known~\cite[Theorem 2]{White1980} that, for
\begin{align*}
(\alpha^\star,\beta^\star)&=\arg\min_{\alpha,\beta} \E[(V-\alpha-\beta U)^2]=\left (\gamma-\mu\frac{\cov(U,V)}{\var(U)},\frac{\cov(U,V)}{\var(U)}\right)
\end{align*}
we have
\begin{align}
\frac{1}{n-2}\sum_{i=1}^n \delta_i^2\to \E[(V-\alpha^\star-\beta^\star U)^2]=\var(V)(1-\rho_{U,V}^2)~~\text{ a.s. as }n\to\infty.
\end{align}
Additionally, by the strong law of large numbers,
\begin{align}
\frac{n/N_n}{n-1}\sum_{i=1}^n (V_i-\overline V)^2\to r\var(V)~~\text{ a.s. as }n\to\infty.
\end{align}
Hence,
$$
\widehat \nu^\cvpp\to (1-r)\var(V)(1-\rho_{U,V}^2)+r\var(V)=\var(V)(1-(1-r)\rho_{U,V}^2)
$$
almost surely as $n\to\infty$.

\subsection{Proof of \cref{thm:m}}

We apply \cref{thm:AsympCSlocalprioriid} to the \iid sequence $(\loss_{\theta}'(\widetilde X_i,f(\widetilde X_i)))_{i\geq 1}$, to obtain an \AsympCS $(\widetilde{\mathcal{R}}_{\delta,\theta,i})_{i\geq 1}$ for $m_\theta$ with approximation rate $1/\sqrt{i\log{i}}$. The subsequence $(\mathcal R_{\delta,\theta,n})_{n\geq 1}$ with $\mathcal R_{\delta,\theta,n}=\widetilde{\mathcal R}_{\delta,\theta,N_n}$ is also an \AsympCS for $m_\theta$ with approximation rate $1/\sqrt{n\log{n}}$. Asymptotic Type-I error control follows directly from \cref{thm:typeIerror}.

%\looseness=-1

\subsection{Proof of \cref{thm:delta}}

In the PPI case, the proof follows from a direct application of \cref{thm:AsympCSlocalprioriid} (non-assisted) or \cref{thm:AsympCSglobalprioriid} (Bayes-assisted) to the sequence of \iid random variables  $(V_{\theta,i}-U_{\theta,i})_{i=1}^n$. In the PPI\texttt{++} case, it follows from an application of \cref{thm:AsympCSlocalpriornoniid} (non-assisted) or \cref{thm:AsympCSglobalpriornoniid} (Bayes-assisted), together with \cref{prop:strassencvestimator}, to the control variate estimator \eqref{eq:deltahatpowertuning}. Asymptotic Type-I error control follows directly from \cref{thm:typeIerror}.

% !TEX root = ppics_supp.tex

\section{Multivariate \AsympCS}
\label{supp:multivariate}

In this section, we discuss how the results developed in this paper for scalar $\theta$ can be extended to obtain asymptotic confidence regions for $\theta\in\bbR^d$.

We first provide the definitions of a multivariate confidence sequence and of an asymptotic spherical confidence sequence. Let $B(x,r)\subset \bbR^d$ denote the ball of radius $r$ centered at $x$. 
\subsection{Definitions}

\begin{definition}(Confidence Sequence)
\label{def:CSmultivariate}
Let $(\calC_{\alpha,t})_{t\geq 1}$ be a sequence of random subsets of $\bbR^d$. For $\alpha\in(0,1)$,  $(\calC_{\alpha,t})_{t\geq 1}$ is a $1-\alpha$ confidence sequence for a fixed parameter $\mu\in\bbR^d$ if
\[\Pr(\mu\in \calC_{\alpha,t}\text{  ~for all }t\geq 1)\geq 1-\alpha.\]
\end{definition}
We now introduce the notion of an asymptotic spherical confidence sequence, inspired by \citep[Section B.10]{WaudbySmith2024a}.

\begin{definition}(Asymptotic Spherical Confidence Sequence)
\label{def:asympCSmultivariate}
Let $\alpha\in(0,1)$ and $(a_t)_{t\geq 0}$ a real sequence such as $\lim_{t\to\infty} a_t=0$. Let $(\widehat\mu_t)_{t\geq 1}$ be a consistent sequence of estimators of $\mu$. The sequence of random balls $(\calC_{\alpha,t})_{t\geq 1}$, with $\calC_{\alpha,t}=B(\widehat\mu_t,R_t)$ and $R_t>0$, is said to be an \textit{asymptotic spherical confidence sequence} with (little-o) approximation rate $a_t$, if there exists a (usually unknown) confidence sequence  $(\calC^\star_{\alpha,t})_{t\geq 1}$, with  $\calC^\star_{\alpha,t}=B(\widehat\mu_t, R_t^\star)$, such that
\[\Pr(\mu\in \calC^\star_{\alpha,t}\text{ for all }t\geq 1)\geq 1-\alpha\]
and
$$
   |R_t-R^\star_t|= o(a_t) \text{~a.s. as~} t\rightarrow+\infty.
$$
\end{definition}

\subsection{Nonasymptotic Bayes-assisted CS for \iid Gaussian random vectors}
\label{supp:multivariateCDiidGaussian}

Let $Y_1,Y_2,\ldots,$ be \iid Gaussian random vectors with mean $\mu\in\bbR^d$ and known $d$-by-$d$ positive definite covariance matrix $\Sigma$. Let $\prior$ be some prior on $\Sigma^{-1/2}\mu$ and define $\eta_t(z)=\int \Normal(z;\zeta,I_d/t) \prior(\zeta)d\zeta$ where $I_d$ denotes the $d$-by-$d$ identity matrix. By the method of mixtures and Ville's inequality (similarly as in \cref{thm:iidgaussian}), the sequence of ellipsoid regions defined by
$$
\calC_{\alpha,t}(\overline Y_t,\Sigma;\pi)=\left \{ \mu\in\bbR^d \mid  \|\Sigma^{-1/2}(\mu-\overline Y_t) \|\leq \frac{1}{\sqrt t} \sqrt{\log\left(\frac{t^d}{(2\pi)^d\alpha^2\eta_t(\Sigma^{-1/2} \overline Y_t)^2}\right)}    \right\}
$$
forms a $(1-\alpha)$ confidence sequence for $\mu$. One could also consider spherical confidence intervals using $\Lambda_{\max}(\Sigma)$, the maximum eigenvalue of $\Sigma$, similarly to what is done in \cite[Section B10]{WaudbySmith2024a} for non-assisted confidence regions. In this case, the corresponding (more conservative) $(1-\alpha)$ confidence sequence for $\mu$ is
\begin{align}
\calC^\mBA_{\alpha,t}(\overline Y_t,\Sigma;\pi)=\left \{ \mu\in\bbR^d \mid  \| \mu-\overline Y_t \| \leq \frac{\sqrt{\Lambda_{\max}(\Sigma)}}{\sqrt t} \sqrt{\log\left(\frac{t^d}{(2\pi)^d\alpha^2\eta_t(\Sigma^{-1/2} \overline Y_t)^2}\right)}    \right\}.
\label{eq:mBACS}
\end{align}

\subsection{Multivariate extension of \cref{thm:AsympCSglobalprioriid,thm:AsympCSglobalpriornoniid}}

To illustrate that our results extend to the multivariate case, we provide multivariate (and tight) versions of \cref{thm:AsympCSglobalpriornoniid} and \cref{thm:AsympCSglobalprioriid}.

\begin{theorem}{(Multivariate version of \cref{thm:AsympCSglobalpriornoniid})}
\label{thm:multiAsympCSgeneral}
Let $(\widehat{\mu}_t)_{t\geq1}$ be a consistent sequence of estimators of $\mu$. Assume that there exists a sequence of \iid Gaussian vectors with mean $\mu$ and positive definite covariance matrix $\Sigma$ such that, a.s. as $t\rightarrow\infty$,
\begin{equation}
% \label{condition1}
\widehat{\mu}_t=\frac{1}{t} \sum_{i=1}^t W_i+\varepsilon_t
\text{~~~where~~~}
\varepsilon_t=o\left(\frac{1}{\sqrt{t \log t}}\right).
\end{equation}
Let $(\widehat\Sigma_t)_{t\geq1}$ be a consistent sequence of estimators of $\Sigma$ such that $\|\widehat\Sigma_t-\Sigma\|=o(1/\log t)$ a.s. where $\|\cdot\|$ is the spectral norm, and $\prior$ be a continuous and proper prior density on $\mathbb{R}^d$. Then, $\calC^\mBA_{\alpha,t}(\widehat{\mu}_t,\widehat\Sigma_t;\prior)$ forms a $(1-\alpha)$--\AsympCS with approximation rate $1/\sqrt{t\log{t}}$.

\end{theorem}

\begin{proof}(\cref{thm:multiAsympCSgeneral})
By hypothesis, we have, almost surely,
\[
\widehat\mu_t =\overline{W}_{t}+\varepsilon_{t},
\]
where $\varepsilon_{t}=o\left(\frac{1}{\sqrt{t\log t}}\right)$ and $\overline W_t=\frac{1}{t}\sum_{i=1}^t W_i$. As stated in \cref{supp:multivariateCDiidGaussian}, the sequence of balls $\calC^\mBA_{\alpha,t}(\overline W_t,\Sigma;\prior)=\calC^\mBA_{\alpha,t}(\widehat\mu_t-\varepsilon_t,\Sigma;\prior)$ forms an exact $(1-\alpha)$ CS for $\mu$. Let
\begin{align*}
R_{t}^{\star} &  =\frac{\sqrt{\Lambda_{\max}(\Sigma)}}{\sqrt t} \sqrt{\log\left(\frac{t^d}{(2\pi)^d\alpha^2\eta_t(\Sigma^{-1/2}(\widehat\mu_t-\varepsilon_t))^2}\right)} +\|\varepsilon_t\|.
\end{align*}
As $B\left(\widehat\mu_t,R_t^\star\right)\supseteq \calC^\mBA_{\alpha,t}(\widehat\mu_t-\varepsilon_t,\Sigma;\prior)$, the sequence of random balls $B\left(\widehat\mu_t,R_t^\star\right)$ also forms an exact $(1-\alpha)$ CS for $\mu$. 
Define $\calC^\mBA_{\alpha,t}(\widehat\mu_t,\widehat\Sigma_t;\prior)=B\left(\widehat\mu_t,R_t\right)$, where
\[
R_{t} =\frac{\sqrt{\Lambda_{\max}(\widehat\Sigma_t)}}{\sqrt t} \sqrt{\log\left(\frac{t^d}{(2\pi)^d\alpha^2\eta_t(\widehat\Sigma_t^{-1/2} \widehat\mu_t)^2}\right)}.
\]
By the Courant-Fischer theorem, we have,
$$
|\Lambda_{\max}\left(\widehat{\Sigma}_t\right)-\Lambda_{\max}(\Sigma)| \leq \|\widehat{\Sigma}_t-\Sigma\|=o(1/\log t)\text{ a.s.}
$$
Additionally, by \cref{thm:molifier}, we have
\[
   \eta_t(\widehat\Sigma_t^{-1/2} \widehat\mu_t)  \rightarrow\prior\left(
    \Sigma^{-1/2}\mu\right)  \text{~and~}  \eta_t(\Sigma^{-1/2}(\widehat\mu_t-\varepsilon_t)) \rightarrow\prior\left(
    \Sigma^{-1/2}\mu\right)   \text{~a.s. as~} t\rightarrow\infty.
\]

It follows that
\begin{align*}
R_t-(R_{t}^{\star}-\|\varepsilon_t\|)&=\frac{R_t^2-(R_{t}^{\star}-\|\varepsilon_t\|)^2}{R_t+(R_{t}^{\star}-\|\varepsilon_t\|)}\\
&\sim\frac{(\Lambda_{\max}(\widehat\Sigma_t)-\Lambda_{\max}(\Sigma))\frac{\log t^d}{t}}{\sqrt{\Lambda_{\max}(\Sigma)\frac{\log t^d}{t}}}\\
&=o(1/\sqrt{t\log t})
\end{align*}
a.s. Then $|R_t-R^\star_t|= o(1/\sqrt{t\log t})$ and   so $\calC^\mBA_{\alpha,t}(\widehat{\mu}_t,\widehat\Sigma_t;\prior)$ is a $(1-\alpha)$--\AsympCS with approximation rate $1/\sqrt{t\log{t}}$.
\end{proof}

\begin{theorem}{(Multivariate version of \cref{thm:AsympCSglobalprioriid})}
\label{thm:multiAsympCSiid}
Let $(Y_t)_{t\geq 1}$ be a sequence of \iid random vectors in $\mathbb{R}^d$ with mean $\mu$ and such that $\mathbb{E}\| Y_1\|^{2+\delta}<\infty$ for some $\delta>0$.
Then, $\calC^\mBA_{\alpha,t}(\overline{Y}_t,\widehat\Sigma_t;\prior)$ is a $(1-\alpha)$--\AsympCS  with approximation rate $1/\sqrt{t\log{t}}$, where $\overline{Y}_t$ is the sample mean and $\widehat\Sigma_t$ the sample covariance.
\end{theorem}

\begin{proof}(\cref{thm:multiAsympCSiid})
By the strong law of large numbers, $\overline{Y}_t$ and $\widehat\Sigma_t$ are consistent estimators of $\mu$ and $\Sigma$, respectively. By the multivariate KMT coupling due to \citet{Einmahl1987} (see \cref{thm:multiKMT}), there exists a sequence of \iid Gaussian random vectors $(W_i)_{i\geq 1}$ with mean $\mu$ and covariance matrix $\Sigma$ such that
    $$
        \overline{Y}_t=\frac{1}{t} \sum_{i=1}^t W_i+\varepsilon_t
        \text{~~~where~~~}
        \varepsilon_t=o\left(\frac{1}{t^{1-1/(2+\delta)}}\right) =o\left(\frac{1}{\sqrt{t\log t}}\right).
    $$
    The result then follows from \cref{thm:multiAsympCSgeneral}.
\end{proof}

All other results can be extended to the multivariate case in a similar manner. In the case of \cref{thm:AsympCSlocalprioriid} and \cref{thm:AsympCSlocalpriornoniid}, we require a multivariate, non-assisted, exact confidence sequence for \iid Gaussian random variables with known variance. For any $\rho > 0$, \citet[Equation (29)]{WaudbySmith2024b} propose to use
\[
\calC^{\mNA}_{\alpha,t}:=\left\{\mu \in \mathbb{R}^d:\left\|\overline{Y}_t-\mu\right\|<\sqrt{\frac{\Lambda_{\max}(\widehat\Sigma_t)\cdot 9 d}{2} \cdot \frac{1+t \rho^2}{t^2 \rho^2} \cdot\left[2+\log \left(\frac{\sqrt{1+t \rho^2}}{\alpha}\right)\right]}\right\},
\]
although alternative constructions are possible. To extend our results on control variates and PPI to the multivariate setting, the proofs can be adapted accordingly. In doing so, we will require multivariate versions of the Marcinkiewicz-Zygmund strong law of large numbers (see \citet[Theorem 7.9]{Ledoux1991}) and of the law of the iterated logarithm (see \citet[Corollary 1]{Koval2002}).

% !TEX root = ppics_supp.tex

\section{Derivations for prediction-powered mean estimation}
\label{supp:mean}

Throughout the main text, we report expressions of quantities related to the construction of prediction-powered \AsympCS for mean estimation.
Here, we explicitly derive those expressions.

The convex loss associated with the estimand $\theta^\star = \mathbb{E}[Y]$ is the squared loss $\loss_\theta(x, y) = (\theta - y)^2/2$, whose subgradient with respect to $\theta$ is given by $\loss_\theta'(x, y) = \theta - y$.
As a result of this, the measure of fit $m_\theta$ takes the form
\begin{align}
    m_\theta = \theta - \mathbb{E}\left[f(X)\right],
\end{align}
whereas the rectifier $\Delta_\theta$ is given by
\begin{align}
    \Delta_\theta = \mathbb{E}\left[f(X) - Y\right].
\end{align}
In particular, notice that, in the case of mean estimation, the rectifier is independent of $\theta$, i.e.~$\Delta_\theta = \Delta_0$ for all $\theta$.

As discussed in \cref{sec:ppi}, PPI uses the sample mean as an estimator of $m_\theta$, which in this case is given by
\begin{align}
    \widehat m_{\theta,n} = \theta - \frac{1}{N_n}\sum_{j=1}^{N_n} f(\widetilde X_j).
\end{align}
For $\Delta_0$, either the PPI estimator $\widehat\Delta^\PP_{0, n}$ \eqref{eq:deltasamplemean} or the PPI\texttt{++} estimator $\widehat\Delta^\PPpp_{0, n}$ \eqref{eq:deltahatpowertuning} may be used.
In the case of mean estimation, these are given by
\begin{align}
    \widehat\Delta^\PP_{0, n} &= \frac{1}{n} \sum_{i=1}^n \left(f(X_i) - Y_i\right), \\
    \widehat\Delta^\PPpp_{0, n} &= \widehat\Delta^\PP_{n} - (\widehat{\lambda}_{\theta,n} - 1)\left(\frac{1}{N_n}\sum_{j=1}^{N_n} f(\widetilde X_j) - \frac{1}{n}\sum_{i=1}^n f(X_i)\right) \\
    &= \frac{1}{n} \sum_{i=1}^n \left(\widehat\lambda_{\theta,n} f(X_i) - Y_i\right) - (\widehat{\lambda}_{\theta,n} - 1)\frac{1}{N_n}\sum_{j=1}^{N_n} f(\widetilde X_j),
\end{align}
where
\begin{equation}
    \widehat\lambda_{\theta,n} = \frac{\widehat\cov\left((\loss_{\theta}'(X_i,Y_i),\loss_{\theta}'(X_i,f(X_i)))_{i=1}^{n}\right)}{\widehat\var\left((\loss_{\theta}'(X_i,f(X_i)))_{i=1}^{n}\right)} = \frac{\widehat\cov((Y_i, f(X_i))_{i=1}^n)}{\widehat\var((f(X_i))_{i=1}^n)}.
\end{equation}
Again, the control-variate parameter $\widehat{\lambda}_{\theta,n}$ does not depend on $\theta$, i.e.~$\widehat{\lambda}_{\theta,n} = \widehat{\lambda}_{0,n}$ for all $\theta$.

Given $\widehat m_{\theta,n}$ and an estimator $\widehat\Delta_{0,n}$ of $\Delta_0$, the associated prediction-powered estimator of $\theta^\star$ is found by solving, in $\theta$, the equation
\begin{equation*}
    \widehat g_{\theta,n} = \widehat m_{\theta,n} + \widehat\Delta_{0,n}.
\end{equation*}
For the two estimators of $\Delta_0$ discussed above, this quantity takes the form
\begin{align}
    \widehat g^\PP_{\theta,n} &= \theta - \frac{1}{N_n}\sum_{j=1}^{N_n} f(\widetilde X_j) + \frac{1}{n} \sum_{i=1}^n \left(f(X_i) - Y_i\right) \\
    &= \theta - \frac{1}{n} \sum_{i=1}^n Y_i + \left(\frac{1}{n} \sum_{i=1}^n f(X_i) - \frac{1}{N_n}\sum_{j=1}^{N_n} f(\widetilde X_j)\right), \\
    \widehat g^\PPpp_{\theta,n} &= \theta - \widehat{\lambda}_{0,n}\frac{1}{N_n}\sum_{j=1}^{N_n} f(\widetilde X_j) + \frac{1}{n} \sum_{i=1}^n \left(\widehat\lambda_{0,n} f(X_i) - Y_i\right) \\
    &= \theta - \frac{1}{n} \sum_{i=1}^n Y_i + \widehat\lambda_{0,n} \left(\frac{1}{n} \sum_{i=1}^n f(X_i) - \frac{1}{N_n}\sum_{j=1}^{N_n} f(\widetilde X_j)\right),
\end{align}
whose zeroes are given by
\begin{align}
    \widehat\theta_n^\PP &= \frac{1}{n}\sum_{i=1}^{n}Y_i - \left(\frac{1}{n} \sum_{i=1}^n f(X_i) - \frac{1}{N_n}\sum_{j=1}^{N_n} f(\widetilde X_j)\right), \\
    \widehat\theta_n^\PPpp &= \frac{1}{n}\sum_{i=1}^{n}Y_i - \widehat{\lambda}_{0,n}\left(\frac{1}{n} \sum_{i=1}^n f(X_i) - \frac{1}{N_n}\sum_{j=1}^{N_n} f(\widetilde X_j)\right),
\end{align}
which match the expressions in \cref{eq:thetahatppi,eq:thetahatppipp}, respectively.

As discussed in \cref{sec:introduction}, a prediction-powered $(1 - \alpha)$ \AsympCS $(\calC_{\alpha,n}^\avpp)_{n\geq 1}$ for $\theta$ is defined through \cref{eq:ppconfidenceinterval} by first constructing a valid \AsympCS $(\calC^g_{\alpha,\theta,n})_{n \geq 1}$ for $g_\theta$.

\cref{sec:avppi} defines valid \AsympCS for $g_\theta$ that incorporate no prior information.
In particular, $\calC^g_{\alpha,\theta,n}$ is constructed as
\begin{align}
    \calC^{g}_{\alpha,\theta,n} &= \calC^\NA_{\alpha,t}(\widehat g_{\theta,n},\widehat\sigma^g_{\theta,n};\rho) \\
    &= \left[ \widehat g_{\theta,n} \pm \frac{\widehat\sigma^g_{\theta,n}}{\sqrt{n}} \sqrt{\left(1+\frac{1}{n\rho^2}\right)\log\left(\frac{n\rho^2+1}{\alpha^2}\right)} \right ],
\end{align}
where $\calC^\NA_{\alpha,n}$ is defined in \cref{thm:AsympCSlocalprioriid}, $\widehat g_{\theta,n}$ is either the PPI estimator $\widehat g^\PP_{\theta,n}$ or the PPI\texttt{++} estimator $\widehat g^\PPpp_{\theta,n}$, and $(\widehat\sigma^g_{\theta,n})^2$ is the corresponding variance estimator, as defined in \cref{prop:avppi_Cg}. More specifically, under the squared loss, for $\widehat g^\PP_{\theta,n}$, we have
\begin{align}
    (\widehat\sigma^g_{\theta,n})^2 &=\frac{1}{n-2}\sum_{i=1}^n \left(Y_i - f(X_i) - \frac{1}{n} \sum_{k=1}^n \left(Y_k - f(X_k)\right)\right)^2 \\
    &\qquad\qquad+\frac{n/N_n}{N_n-1}\sum_{j=1}^{N_n} \left(f(\widetilde X_j) - \frac{1}{N_n} \sum_{k=1}^{N_n} f(\widetilde X_k)\right)^2 ,
\end{align}
while, for $\widehat g^\PPpp_{\theta,n}$, we have
\begin{align}
    (\widehat\sigma^g_{\theta,n})^2 &=\frac{1-n/N_n}{n-2}\sum_{i=1}^n \left(Y_i -  \widehat{\lambda}_{0,n} f(X_i) - \frac{1}{n} \sum_{k=1}^n \left(Y_k -  \widehat{\lambda}_{0,n} f(X_k)\right)\right)^2 \\
    &\qquad\qquad + \frac{n/N_n}{n-1}\sum_{i=1}^n \left(Y_i - \frac{1}{n}\sum_{k=1}^n Y_k\right)^2.
\end{align}
Given the specific form of $\widehat g_{\theta,n}$ under the squared loss, $\calC_{\alpha,n}^\avpp$ can be explicitly expressed as
\begin{align}
    \calC_{\alpha,n}^\avpp &= \left\{\theta \mid 0 \in \calC^{g}_{\alpha,\theta,n}\right\} \\
    &= \left[\widehat{\theta}_n \pm \frac{\widehat\sigma^g_{0,n}}{\sqrt{n}} \sqrt{\left(1+\frac{1}{n\rho^2}\right)\log\left(\frac{n\rho^2+1}{\alpha^2}\right)} \right],
\end{align}
which is an interval, and where $\widehat{\theta}_n$ is either the PPI estimator $\widehat\theta_n^\PP$ or the PPI\texttt{++} estimator $\widehat\theta_n^\PPpp$.

Similarly, \cref{sec:avbappi} defines valid \AsympCS for $g_\theta$ that incorporate prior information by means of a zero-mean prior $\pi$ on $\Delta_\theta$.
In particular, for $\delta \in (0, \alpha)$, \cref{thm:m} first construct a standard $(1 - \delta)$ \AsympCS $\mathcal R_{\delta,\theta,n}$ for $m_\theta$, which in the case of mean estimation takes the form
\begin{align}
    \mathcal R_{\delta,\theta,n} &= \calC^\NA_{\delta,n}(\widehat m_{\theta,n},\widehat\sigma^f_{\theta,n};\rho) \\
    &= \left[\theta - \frac{1}{N_n}\sum_{j=1}^{N_n} f(\widetilde X_j) \pm \frac{\widehat\sigma^f_{\theta,n}}{\sqrt{N_n}} \sqrt{\left(1+\frac{1}{N_n\rho^2}\right)\log\left(\frac{N_n\rho^2+1}{\delta^2}\right)} \right ],
\end{align}
where $(\widehat\sigma^f_{\theta,n})^2$ is the sample variance of $(\loss_{\theta}'(\widetilde X_i,f(\widetilde X_i)))_{i=1}^{N_n}$, that is
\begin{align}
    (\widehat\sigma^f_{\theta,n})^2 &= \widehat\var((\loss_{\theta}'(\widetilde X_i,f(\widetilde X_i)))_{i=1}^{N_n}) \\
    &= \frac{1}{N_n - 1} \sum_{j=1}^{N_n} \left(f(\widetilde X_j) - \frac{1}{N_n}\sum_{k=1}^{N_n} f(\widetilde X_k)\right)^2.
\end{align}
Next, \cref{thm:delta} constructs a Bayes-assisted $(1 - (\alpha - \delta))$ \AsympCS $\mathcal T_{\alpha - \delta,\theta,n}$ for $\Delta_\theta$, which under the squared loss takes the form
\begin{align}
    \mathcal T_{\alpha - \delta,\theta,n} &= \calC_{\alpha - \delta, n}^\BA(\widehat \Delta_{0,n},\widehat\sigma^\Delta_{\theta,n}; \pi) \\
    &= \left[ \widehat \Delta_{0,n} \pm \frac{\widehat\sigma^\Delta_{\theta,n}}{\sqrt n} \sqrt{\log\left(\frac{n(2\pi(\alpha - \delta)^2)^{-1}}{\eta_n(\widehat \Delta_{0,n}/\widehat\sigma^\Delta_{\theta,n})^2}\right)}\right],
\end{align}
where $\calC^\BA_{\alpha-\delta,n}$ and $\eta_n$ are defined in \cref{thm:AsympCSglobalprioriid}, $\widehat \Delta_{0,n}$ is either the PPI estimator $\widehat\Delta^\PP_{0, n}$ or the PPI\texttt{++} estimator $\widehat\Delta^\PPpp_{0, n}$, and $(\widehat\sigma^\Delta_{\theta,n})^2$ is the corresponding variance estimator, as defined in \cref{thm:delta}.
In the case of mean estimation, $(\widehat\sigma^\Delta_{\theta,n})^2$ takes the form
\begin{align}
    (\widehat\sigma^\Delta_{\theta,n})^2 = \frac{1}{n - 1}\sum_{i=1}^n \left(Y_i - f(X_i) - \frac{1}{n}\sum_{k=1}^n\left(Y_k - f(X_k)\right)\right)^2,
\end{align}
for PPI, and is given by
\begin{align}
    (\widehat\sigma^\Delta_{\theta,n})^2 &= \frac{1-n/N_n}{n-2}\sum_{i=1}^n \left(Y_i - \widehat{\lambda}_{0,n} f(X_i)  - \frac{1}{n}\sum_{k=1}^n\left(Y_k - \widehat{\lambda}_{0,n} f(X_k)\right)\right)^2 \\
    &\qquad\qquad + \frac{n/N_n}{n-1} \sum_{i=1}^n \left(Y_i - f(X_i) - \frac{1}{n}\sum_{k=1}^n\left(Y_k - f(X_k)\right)\right)^2
\end{align}
for PPI\texttt{++}.
Finally, $\mathcal T_{\alpha - \delta,\theta,n}$ and $\mathcal R_{\delta,\theta,n}$ are combined via a Minkowski sum to construct a valid $(1 - \alpha)$ \AsympCS for $g_\theta$ as
\begin{align}
    \calC^{g}_{\alpha,\theta,n} &= \mathcal T_{\alpha - \delta,\theta,n} + \mathcal R_{\delta,\theta,n} \\
    &= \left[\widehat g_{\theta,n} \pm \left\{ \frac{\widehat\sigma^\Delta_{\theta,n}}{\sqrt n} \sqrt{\log\left(\frac{n(2\pi(\alpha - \delta)^2)^{-1}}{\eta_n(\widehat \Delta_{0,n}/\widehat\sigma^\Delta_{\theta,n})^2}\right)} + \frac{\widehat\sigma^f_{\theta,n}}{\sqrt{N_n}} \sqrt{\left(1+\frac{1}{N_n\rho^2}\right)\log\left(\frac{N_n\rho^2+1}{\delta^2}\right)}\right\}\right],
\end{align}
where $\widehat g_{\theta,n}$ is either the PPI estimator $\widehat g^\PP_{\theta,n}$ or the PPI\texttt{++} estimator $\widehat g^\PPpp_{\theta,n}$.
As above, the form of $\widehat g_{\theta,n}$ for mean estimation allows expressing $\calC_{\alpha,n}^\avpp$ explicitly as
\begin{align}
    \calC_{\alpha,n}^\avpp &= \left\{\theta \mid 0 \in \calC^{g}_{\alpha,\theta,n}\right\} \\
    &= \left[\widehat{\theta}_n \pm \left\{\frac{\widehat\sigma^\Delta_{\theta,n}}{\sqrt n} \sqrt{\log\left(\frac{n(2\pi(\alpha - \delta)^2)^{-1}}{\eta_n(\widehat \Delta_{0,n}/\widehat\sigma^\Delta_{\theta,n})^2}\right)} + \frac{\widehat\sigma^f_{\theta,n}}{\sqrt{N_n}} \sqrt{\left(1+\frac{1}{N_n\rho^2}\right)\log\left(\frac{N_n\rho^2+1}{\delta^2}\right)}\right\} \right],
\end{align}
which matches the expression in \cref{eq:avbappi_Ctheta_squared_loss}, and where $\widehat{\theta}_n$ is either the PPI estimator $\widehat\theta_n^\PP$ or the PPI\texttt{++} estimator $\widehat\theta_n^\PPpp$.

% !TEX root = ppics_supp.tex

\section{Experimental details}
\label{supp:experimental_details}

\subsection{Implementation}
Code implementing our method is written in Python and made available at \url{https://github.com/stefanocortinovis/ppi-cs}.
All experiments were run locally on an Apple Silicon M4 Pro CPU with 24GB of memory.

\subsection{Datasets}\label{supp:datasets}
Here we briefly describe each dataset used for the real data experiments in \cref{sec:real_data}.
The \textsc{flights} dataset was downloaded from Kaggle\footnote{\url{https://www.kaggle.com/datasets/shubhambathwal/flight-price-prediction}}, while all the others are available as part of the \texttt{ppi-python} package\footnote{\url{https://pypi.org/project/ppi-python/}}.

\paragraph{Flights.}
For each of $103333$ economy class flight tickets, the \textsc{flights} dataset reports the ticket price ($Y_i \in \mathbb{R}$), as well as the prediction of a gradient-boosted tree for $Y_i$ ($f(X_i) \in \mathbb{R}$).
The goal is to estimate the average price of a flight, i.e.~$\theta^\star = \mathbb{E}[Y] \in \mathbb{R}$.

\paragraph{Forest.}
For each of $1596$ parcels of land in the Amazon rainforest \citep{bullock2020satellite}, the \textsc{forest} dataset reports whether the parcel has been subject to deforestation ($Y_i \in \{0, 1\}$), as well as the prediction of a gradient-boosted tree model for the probability of $Y_i$ being equal to one ($f(X_i) \in [0, 1]$).
The goal is to estimate the fraction of Amazon rainforest lost to deforestation, i.e.~$\theta^\star = \mathbb{E}[Y] \in [0, 1]$.

\paragraph{Galaxies.}
For each of $16743$ images from the Galaxy Zoo 2 initiative \citep{willett2013galaxy}, the \textsc{galaxies} dataset reports whether the galaxy has spiral arms ($Y_i \in \{0, 1\}$), as well as the prediction of a ResNet50 model \citep{he2016deep} for the probability of $Y_i$ being equal to one ($f(X_i) \in [0, 1]$).
The goal is to estimate the fraction of galaxies with spiral arms, i.e.~$\theta^\star = \mathbb{E}[Y] \in [0, 1]$.

\paragraph{Census.}
For each of $380091$ individuals from the 2019 California census, the \textsc{census} dataset reports the individual's age ($X_i \in \mathbb{R}$) and yearly income ($Y_i \in \mathbb{R}$), as well as the prediction of a gradient-boosted tree model trained on the previous year's data for $Y_i$ ($f(X_i) \in \mathbb{R}$).
The goal is to estimate the ordinary least squares (OLS) regression coefficient when regressing income on age.
We preprocess the data by excluding non-positive incomes ($Y_i \leq 0$ or $f(X_i) \leq 0$), and applying a log-transformation to both the response $Y_i$ and the prediction $f(X_i)$.
This results in a dataset of $268118$ individuals.

\paragraph{Healthcare.}
For each of $318215$ individuals from the 2019 California census, the \textsc{healthcare} dataset reports the individual's yearly income ($X_i \in \mathbb{R}$) and whether they have health insurance ($Y_i \in \{0, 1\}$), as well as the prediction of a gradient-boosted tree model trained on the previous year's data for the probability of $Y_i$ being equal to one ($f(X_i) \in [0, 1]$).
The goal is to estimate the logistic regression coefficient when regressing health insurance status on income.
As above, we preprocess the data by excluding non-positive incomes ($X_i \leq 0$), and applying a log-transformation to the covariate $X_i$.
This results in a dataset of $270214$ individuals.

\paragraph{Genes.}
For each of $61150$ gene promoter sequences \citep{Vaishnav2022}, the \textsc{genes} dataset reports the expression level of the gene induced by the promoter ($Y_i \in \mathbb{R}$), as well as the prediction of a transformer model for $Y_i$ ($f(X_i) \in \mathbb{R}$).
The goal is to estimate the median expression level across sequences.
We preprocess the data by applying a log-transformation to the response $Y_i$.

\subsection{Predictor performance}
\Cref{tab:predictor_performance} reports the performance of the predictors used for each real data dataset above, measured in terms of normalised root mean squared error (NRMSE) for regression (R) tasks (\textsc{flights}, \textsc{census}, \textsc{genes}) and cross-entropy (CE) for the binary classification (C) tasks (\textsc{forest}, \textsc{galaxies}, \textsc{healthcare}).
\begin{table}[h]
    \caption{Predictor performance on real data datasets.}
    \centering
    \begin{tabular}{lcccccc}
        \toprule
        Dataset & Flights & Forest & Galaxies & Census & Healthcare & Genes \\
        \midrule
        Task & R & C & C & R & C & R \\
        Performance & $0.20$ & $0.31$ & $0.29$ & $0.11$ & $0.36$ & $0.33$ \\
        \bottomrule
    \end{tabular}
    \label{tab:predictor_performance}
\end{table}
While we report these for completeness, we emphasise that non-assisted PPI improves over classical inference in the presence of correlation between the predictions and the true labels, regardless of the absolute predictive performance (see e.g.~\cref{fig:simulation_biased}).
On the other hand, while knowledge on the predictive performance can be used to choose a suitable prior for Bayes-assisted PPI, the latter is placed on the rectifier $\Delta_\theta$, which depends on the downstream inference task, thereby making the relationship between predictive performance and efficiency gains less direct.

\subsection{\AsympCS hyperparameters}\label{supp:hyperparameters}
Here we discuss the hyperparameters of the PPI and PPI\texttt{++} \AsympCS procedures defined in \cref{sec:abppi}.

The non-assisted prediction-powered \AsympCS discussed in \cref{sec:avppi} requires the specification of the parameter $\rho$ for \cref{eq:NonBayesianConfidenceSequence} in \cref{thm:AsympCSlocalprioriid}.
As mentioned at the end of \cref{sec:background}, $\rho$ can be chosen so as to minimise the width of the interval at a specified time.
In particular, as shown in \citet[Appendix B.2]{WaudbySmith2024b}, setting
\begin{equation}
    \rho = \sqrt{\frac{-W_{-1}(-\alpha^2 \exp(-1)) - 1}{t^\star}},
\end{equation}
where $W_{-1}$ is the lower branch of the Lambert $W$ function, minimises the width of the interval at time $t^\star$.

The Bayes-assisted prediction-powered \AsympCS discussed in \cref{sec:avbappi} requires the specification of both the parameter $\rho$ used for the non-assisted \AsympCS for the measure of fit $m_\theta$ (\cref{thm:m}) and the scale parameter $\tau$ of the prior $\pi$ for the Bayes-assisted \AsympCS for the rectifier $\Delta_\theta$ (\cref{thm:delta}).
While the former can be chosen as above, the same approach does not work for the latter, as the prior scale that minimises the width of the Bayes-assisted interval at a specified time $t$ depends on the observed value of $\overline Y_t / \widehat\sigma_t$ in \cref{eq:BayesianConfidenceSequence} of \cref{thm:AsympCSglobalprioriid}.
Instead, we propose the following heuristic for choosing $\tau$.
If $Z_1,Z_2,\ldots | \mu \iidsim \mathcal{N}(\mu, 1)$, then the posterior mean $\mathbb{E}[\mu | Z_1,\dots,Z_t]$ after $t$ observations under a Gaussian prior $\mu \sim \mathcal{N}(\mu_0, \tau^2)$ is given by
\begin{equation}
    \mathbb{E}[\mu | Z_1,\dots,Z_t] = \frac{1}{1 + t \tau^2} \mu_0 + \frac{t \tau^2}{1 + t \tau^2} \overline Z_t,
    \label{eq:posterior_mean_heuristic}
\end{equation}
where the two terms on the right-hand side measure the influence of the prior and the data on the posterior mean, respectively.
Noticing that the Gaussian likelihood leading to the posterior mean \eqref{eq:posterior_mean_heuristic} is of the same form as the one implicitly used for the construction of Bayes-assisted \AsympCS (see e.g.~\cref{eq:marginaldensity}), we choose $\tau$ so that the prior and the data have the same influence on the posterior mean \eqref{eq:posterior_mean_heuristic} at time $t^\star$, i.e.
\begin{equation}
    \tau = \frac{1}{\sqrt{t^\star}}.
\end{equation}

\cref{supp:addexp} reports the hyperparameter value used for each experiment in terms of $t^\star$.
Notice that, as discussed in \cref{sec:experiments}, the initial $N_n$ is set large enough to rule out any uncertainty on the measure of fit $m_\theta$.
As a result of this, the only hyperparameter that needs to be chosen for the Bayes-assisted procedure is the prior scale $\tau$.

% !TEX root = ppics_supp.tex

\section{Additional experimental results}
\label{supp:addexp}

Additional experimental results to complement \cref{sec:experiments} are presented here.
Legend names are as in \cref{sec:experiments}.

\subsection{Synthetic data}\label{supp:synthetic_data}
\subsubsection{Noisy predictions}
For this experiment, we set $t^\star = 500$ (see \cref{supp:hyperparameters}) for all methods.

\begin{figure}[h]
    \centering
    \includegraphics[width=\textwidth]{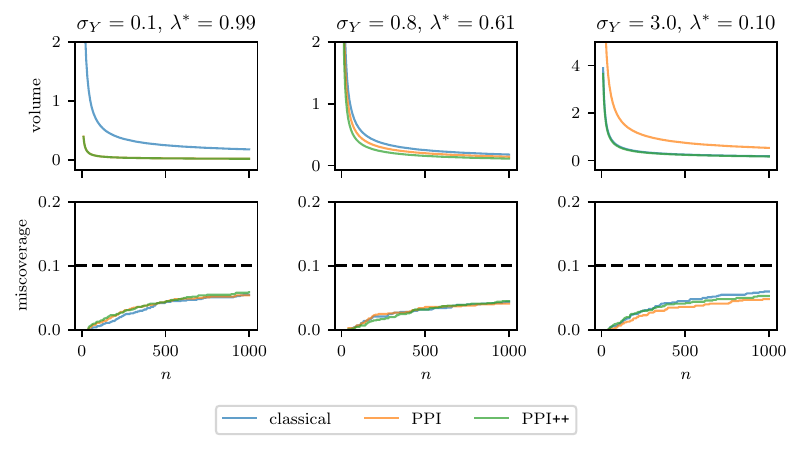}
    \caption{
        Noisy predictions study. The left, middle and right panels show average interval volume and cumulative miscoverage rate over $1000$ repetitions for noise levels $\sigma_Y = 0.1, 0.8, 3.0$.
    }
    \label{fig:simulation_noisy_coverage}
\end{figure}

\cref{fig:simulation_noisy_coverage} reports the average cumulative miscoverage rate for the results shown in \cref{fig:simulation_noisy}.
As desired, the cumulative miscoverage rate lies below the threshold $\alpha$ for all $n$.

\begin{figure}[h]
    \centering
    \includegraphics[width=\textwidth]{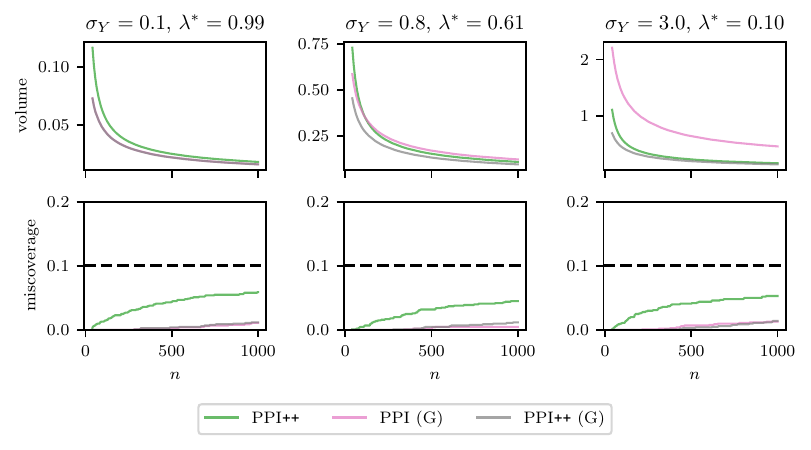}
    \caption{
        Noisy predictions study with Gaussian prior. The left, middle and right panels show average interval volume and cumulative miscoverage rate over $1000$ repetitions for noise levels $\sigma_Y = 0.1, 0.8, 3.0$. Results for non-assisted PPI\texttt{++} are shown for reference.
        }
        \label{fig:simulation_noisy_bayes}
    \end{figure}

\cref{fig:simulation_noisy_bayes} shows the performance of the Bayes-assisted prediction-powered \AsympCS procedures under a Gaussian prior on the noisy predictions experiment in \cref{sec:synthetic_data}.
The results are consistent with those presented in \cref{sec:synthetic_data}.
In particular, also with Bayes-assistance, PPI\texttt{++} easily adapts to increasing noise levels, while standard PPI fails to do so.
Moreover, in this case, Bayes-assisted PPI\texttt{++} outperforms the non-assisted version across all noise levels.
This is due to the fact that, in this experiment, the predictions from $f$, while noisy, are unbiased for all values of $\sigma_Y$.
As a result of this, the zero-mean prior used by the Bayes-assisted procedures is well specified, and any additional shrinkage performed by the latter is beneficial. 

\subsubsection{Biased predictions}
\begin{figure}[h]
    \centering
    \includegraphics[width=\textwidth]{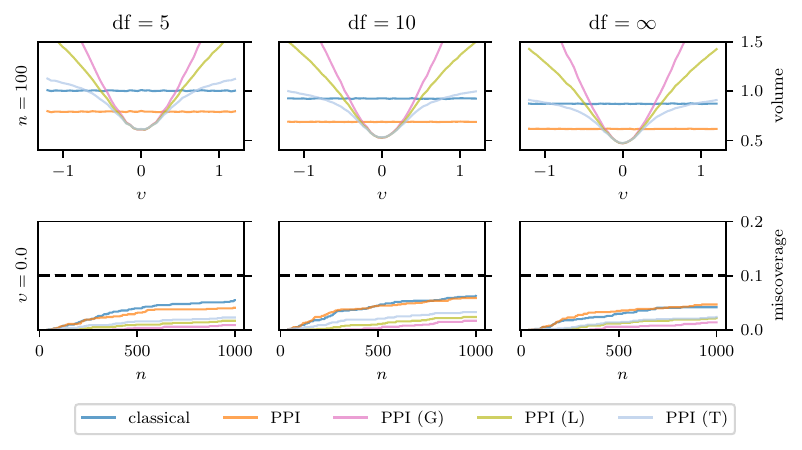}
    \caption{
        Biased predictions study. The left, middle and right panels show average interval volume and cumulative miscoverage rate over $1000$ repetitions for $\mathrm{df} = 5, 10, \infty$.
    }
    \label{fig:simulation_biased_coverage}
\end{figure}

\begin{figure}[h]
    \centering
    \includegraphics[width=\textwidth]{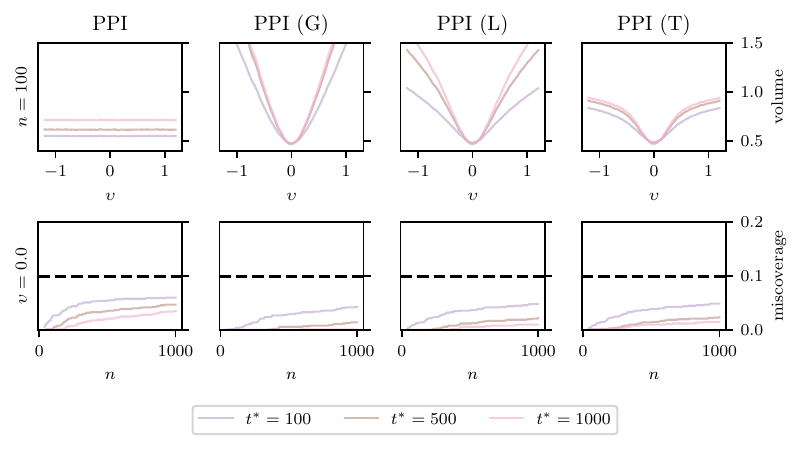}
    \caption{
        Biased predictions study with different hyperparameters. Each column correspond to one of the prediction-powered methods in \cref{fig:simulation_biased} for $t^\star = 100, 500, 1000$.
        The top and bottom column show average interval volume and cumulative miscoverage rate over $1000$ repetitions with $\mathrm{df} = \infty$.
    }
    \label{fig:simulation_biased_scale}
\end{figure}

For this experiment, we set $t^\star = 500$ (see \cref{supp:hyperparameters}) for all methods shown in \cref{fig:simulation_biased,fig:simulation_biased_coverage}.
The values of $t^\star$ used for \cref{fig:simulation_biased_scale} are reported in the figure legend.

\cref{fig:simulation_biased_coverage} reports the average cumulative miscoverage rate for the results shown in \cref{fig:simulation_biased} at $\upsilon = 0$.
As discussed in \cref{sec:synthetic_data}, the cumulative miscoverage rate increases slightly as we decrease $\mathrm{df}$, but remains below the threshold $\alpha$ for all $n$, as desired.

\cref{fig:simulation_biased_scale} repeats the simulation of \cref{fig:simulation_biased_coverage} for different values of $t^\star$, which affects the procedure-specific hyperparameters as discussed in \cref{supp:hyperparameters}.
For non-assisted PPI, $t^\star$ represents the time at which the procedure's interval width is minimised.
Therefore, as expected, increasing $t^\star$ above $100$ leads to larger intervals for $n = 100$.
However, the qualitative behaviour of the non-assisted methods as $\upsilon$ varies is the same across all values of $t^\star$: their volume remain constant across bias levels, reflecting the lack of prior information.
On the other hand, for Bayes-assisted methods, a larger $t^\star$ implies a smaller prior scale $\tau$.
As a result of this, increasing $t^\star$ above $100$ leads to stronger prior influence at $n = 100$.
In particular, a large $t^\star$ results in slightly smaller intervals for $\upsilon \approx 0 $, but larger intervals for $|\upsilon| \gg 0$.
When comparing the results across different priors, the results are consistent with those in \cref{fig:simulation_biased}: the interval volume under heavier-tailed priors, such as the Laplace and Student-t priors, grow at a lower rate with $|\upsilon|$ compared to the Gaussian prior, thereby offering greater robustness to prior misspecification.

\subsubsection{Multivariate biased predictions}
Here, we illustrate the multivariate \AsympCS procedure described in \cref{supp:multivariate} in the context of PPI.
To do this, we study a simple multivariate version of the mean estimation task with biased prediction described in \cref{sec:synthetic_data}.
In particular, for $d = 5$, we sample $d$-dimensional observations $Y_i \iidsim \mathcal{N}(\mathbf{0}, \Sigma)$, where $\Sigma \in \mathbb{R}^{d \times d}$ is a Toeplitz covariance matrix with entries $\Sigma_{ij} = 0.5^{|i-j|}$, and define $Y_i = f(X_i) + \epsilon_i$, where $\epsilon_i \iidsim \mathcal{N}(\mathbf{0}, I_d)$, so that $\theta^\star = \mathbb{E}[Y] = \mathbf{0}$.
Then, we proceed as in \cref{sec:synthetic_data} and define biased predictions $f(X_i) = X_i + \upsilon$, where $\upsilon \in \mathbb{R}$ controls the bias level of the predictor.
In this setup, we compare classical inference, non-assisted PPI, and Bayes-assisted PPI under a Gaussian prior with mean zero and isotropic covariance matrix using the spherical multivariate \AsympCS procedures described in \cref{supp:multivariate}.
The \AsympCS hyperparameters are set using natural extensions of the rules in \cref{supp:hyperparameters} to the multivariate case, with $t^\star = 1000$ for all methods.
\Cref{fig:simulation_biased} shows the average spherical interval volumes of each method as a function of $\upsilon$, which we vary between $-6$ and $6$, at $n \in \{100, 250, 500\}$.
\begin{figure}[h]
    \centering
    \includegraphics[width=\textwidth]{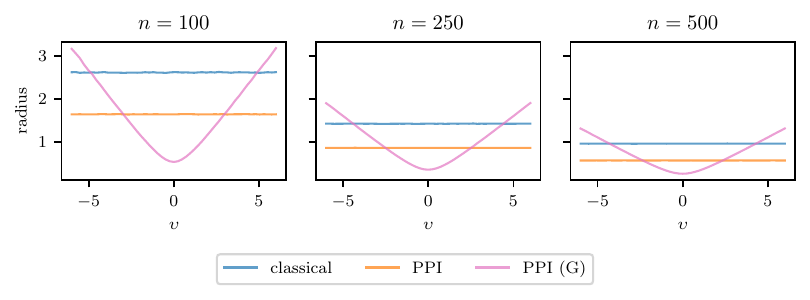}
    \caption{
        Multivariate biased predictions study. The left, middle and right panels show average spherical region radius over $1000$ repetitions for $n = 100, 250, 500$.
    }
    \label{fig:simulation_biased_multivariate}
\end{figure}
The results are consistent with those in \cref{fig:simulation_biased}.
In particular, non-assisted PPI consistently outperforms classical inference, with both methods yielding constant interval radii across bias levels.
On the other hand, Bayes-assisted PPI achieves smaller radii than the other baselines for small values of $\upsilon$, but its radius grows quickly with $|\upsilon|$ as the prior becomes increasingly misspecified.
For this example, we do not report coverage results, as we find that all methods achieve near perfect coverage across the values of $\upsilon$ considered, with cumulative miscoverage rates close to zero, likely due to the conservative spherical construction mentioned in \cref{supp:multivariate}.

\subsection{Real data}\label{supp:real_data}
\subsubsection{Mean estimation}
For each of the mean estimation experiments, we set $t^\star$ (see \cref{supp:hyperparameters}) equal to the largest $n$ considered in the experiment.
In particular, for the \textsc{flights}, \textsc{forest}, and \textsc{galaxies} datasets, we set $t^\star = 10000$, $500$, and $1000$, respectively.

\begin{figure}[h]
    \centering
    \includegraphics[width=\textwidth]{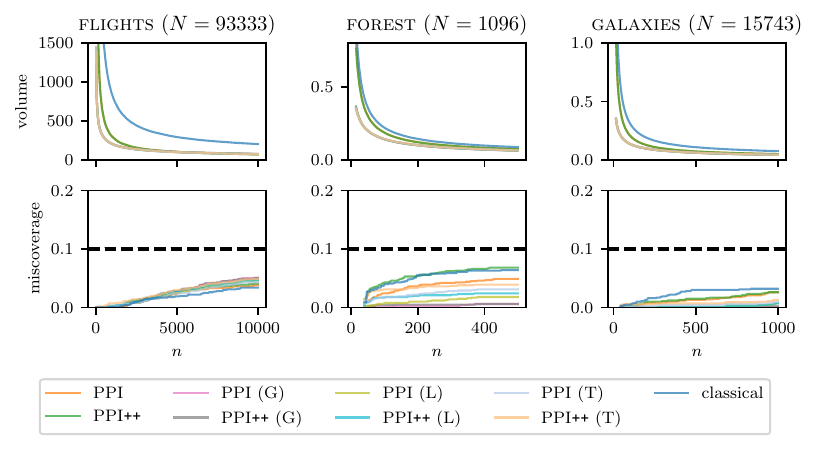}
    \caption{
        Mean estimation. The top and bottom rows show the average interval volume and cumulative miscoverage rate over $1000$ repetitions for the \textsc{flights}, \textsc{forest}, and \textsc{galaxies} datasets.
    }
    \label{fig:real_data_supplementary}
\end{figure}

\cref{fig:real_data_supplementary} adds the results of the standard PPI procedures to the ones shown in \cref{fig:real_data}.
For these experiments, the improvement of PPI\texttt{++} over standard PPI is small, and the results remain consistent with those in \cref{fig:real_data}.
That is, PPI methods consistently improve over classical inference, with Bayes-assisted methods providing an additional efficiency boost for moderate labelled sample sizes.

\subsubsection{Other estimation tasks}
The estimation tasks considered here involve linear regression (\textsc{census} dataset), logistic regression (\textsc{healthcare} dataset), and median estimation (\textsc{genes} dataset).
As above, for each estimation task, we set $t^\star = 2000$ (see \cref{supp:hyperparameters}), as that is the largest $n$ considered in all experiments.

For these, \AsympCS procedures relying on classical inference (obtained from \cref{thm:AsympCSlocalprioriid}) and PPI, both non-assisted (\cref{prop:avppi_Cg}) and Bayes-assisted (\cref{eq:avbappi_Cg}), require constructing a grid over $\theta$ through \cref{eq:ppconfidenceinterval}.
To initialise the grid, we use the first $n_0$ labelled data points to compute a preliminary estimate of $\theta^\star$, which we then use to centre the grid.
The same $n_0$ is also used as the starting point to evaluate the \AsympCS procedures and compute their cumulative miscoverage rate reported in the figures below.
We set $n_0 = 100$ for the \textsc{census} and \textsc{healthcare} datasets, and $n_0 = 40$ for the \textsc{genes} dataset.

Furthermore, some priors, including the Student-$t$ prior, require numerical integration to compute the marginal density $\eta_t$ used in \cref{eq:avbappi_Cg}.
As a result, when Bayes-assisted PPI under such priors is used, the computational cost grows significantly when \cref{eq:avbappi_Cg} is evaluated across many $n$ and $\theta$ values simultaneously.
Because of this, we only report results for the Gaussian and Laplace priors, which admit closed-form expressions for $\eta_t$.

\begin{figure}[h]
    \centering
    \includegraphics[width=\textwidth]{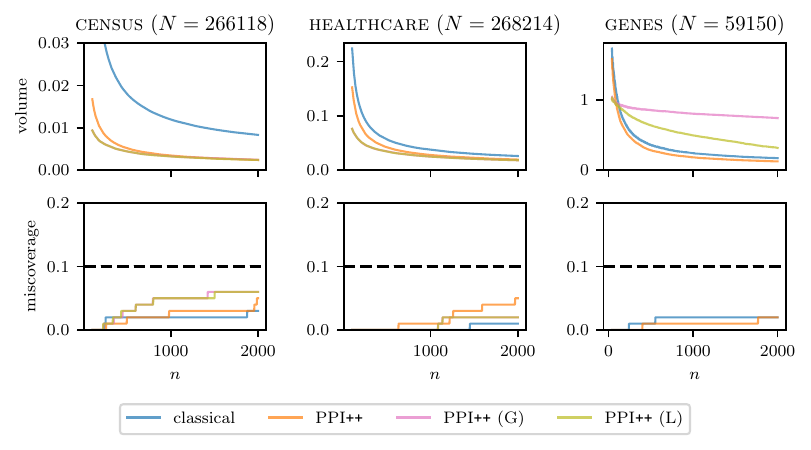}
    \caption{
        Other estimation tasks. The top and bottom rows show the average interval volume and cumulative miscoverage rate over $100$ repetitions for the \textsc{census}, \textsc{healthcare}, and \textsc{genes} datasets.
    }
    \label{fig:other_estimation}
\end{figure}

\cref{fig:other_estimation} compares classical and PPI \AsympCS procedures on the three estimations tasks above in terms of average interval volume and cumulative miscoverage rate as $n$ increases.
As discussed in \cref{sec:real_data}, PPI methods outperform classical inference for the linear and logistic regression tasks, with Bayes-assisted methods further improving efficiency when $n$ is moderate.
For the median estimation task, on the other hand, non-assisted PPI still improves over classical inference, while Bayes-assisted PPI yield larger regions than the other methods due to the higher bias of the predictions in this dataset.
In all cases, coverage remains satisfactory.

\begin{figure}[h]
    \centering
    \includegraphics[width=\textwidth]{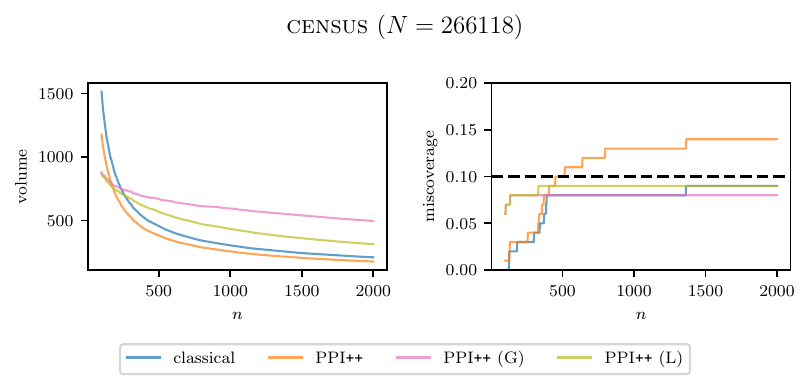}
    \caption{
        Linear regression on the \textsc{census} dataset without preprocessing. The top and bottom rows show the average interval volume and cumulative miscoverage rate over $100$ repetitions.
    }
    \label{fig:census_income_nopreprocessing}
\end{figure}

As discussed in \cref{supp:datasets}, the \textsc{census}, \textsc{healthcare}, and \textsc{genes} datasets are preprocessed by applying a log-transformation to relevant positive skewed variables, as it is commonly done in the literature.
For instance, this is the case for the income variable $Y_i$ in the \textsc{census} dataset.
In practice, such a transformation improves the accuracy of the KMT coupling approximation for a given $n$, essentially lowering the effective labelled sample size necessary to achieve satisfactory coverage.
To see this, we repeat the linear regression experiment on the \textsc{census} dataset without applying any preprocessing to $Y_i$.
As shown in \cref{fig:census_income_nopreprocessing}, the results are strikingly different: all methods yield significantly larger cumulative miscoverage rates compared to the preprocessed case in \cref{fig:other_estimation}, with non-assisted PPI violating the nominal guarantee around $n \approx 500$, in turn invalidating the efficiency comparison.
Without preprocessing, a substantially larger starting labelled sample size $n_0$ is needed before the KMT coupling approximation is accurate enough for satisfactory uniform-time coverage by the \AsympCS procedures.
This example highlights the importance of knowledge of the data distribution when using \AsympCS procedures in practice.
A possible way to obtain such knowledge in practice is to estimate the third moment of the data distribution, if it exists, from a held-out validation set and use a Berry-Esseen-type bound \citep[Theorem~2.1.4]{Vershynin2009} to choose a starting labelled sample size $n_0$ at which the KMT coupling provides a good approximation.

% !TEX root = ppics_supp.tex

\section{Alternative non-assisted \AsympCS}
\label{supp:extendedville}

\subsection{Parameter-free \AsympCS via improper prior}
As discussed in \cref{sec:background}, the non-assisted asymptotic confidence sequence $\calC^\NA_{\alpha,t}(\overline Y_t,\widehat{\sigma}_t;\rho)$ in \cref{eq:NonBayesianConfidenceSequence} approximates the exact CS in \cref{eq:ExactNonBayesianConfidenceSequence} and becomes arbitrarily accurate as in the limit.
This suggests constructing alternative non-assisted \AsympCS by approximating other exact CSs for which adaptations of \cref{thm:AsympCSlocalprioriid,thm:AsympCSlocalpriornoniid} apply.

One example is the parameter-free non-assisted CS of \citet[Corollary 5.9]{Wang2023}.
Define the continuous, strictly decreasing bijection $g\colon[1,\infty)\to(0,1]$ by
\begin{equation*}
    g(x) := 2 \left[1-\Phi \left(\sqrt{\log(x^2)}\right)\right] + 2\sqrt{\log(x^2)}\phi\left(\sqrt{\log(x^2)}\right),
\end{equation*}
with $\Phi$ and $\phi$ the standard normal CDF and PDF, and let $z_\alpha := g^{-1}(\alpha)$, which is well-defined.
The corresponding \AsympCS is given by
\begin{equation}
    \calC^{\NA'}_{\alpha,t}(\overline Y_t,\widehat{\sigma}_t) := \left[\overline{Y}_t \pm \frac{\widehat\sigma_t}{\sqrt{t}} \sqrt{\log\left(t z_\alpha^{2}\right)} \right].
    \label{eq:ImproperConfidenceSequence}
\end{equation}
Notably, for \iid Gaussian observations with known variance, the exact (nonasymptotic) counterpart of \eqref{eq:ImproperConfidenceSequence} is obtained by applying the method of mixtures for extended nonnegative martingales \citep[Def.~3.1]{Wang2023} together with extended Ville's inequality \citep[Theorem 4.1]{Wang2023}, using a non-informative improper prior as the mixing density.

\subsection{Experiments}
We use $\calC^{\NA'}_{\alpha,t}$ from \eqref{eq:ImproperConfidenceSequence} as a parameter-free drop-in replacement for $\calC^\NA_{\alpha,t}$ in the classical and prediction-powered \AsympCS procedures from the main text and repeat some of the experiments from \cref{sec:experiments}.
In figures, runs that use the alternative \AsympCS are annotated (I) for ``improper”.
\begin{figure}[h]
    \centering
    \includegraphics[width=\textwidth]{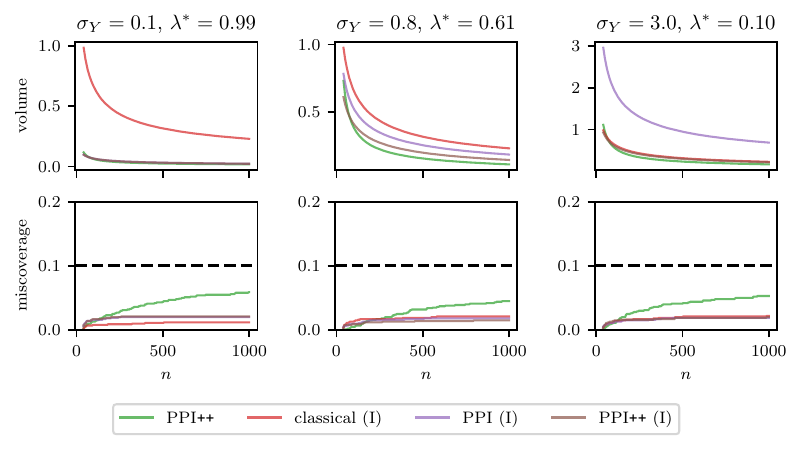}
    \caption{
        Noisy predictions study with alternative \AsympCS. The left, middle and right panels show average interval volume and cumulative miscoverage rate over $1000$ repetitions for noise levels $\sigma_Y = 0.1, 0.8, 3.0$. Results for non-assisted PPI\texttt{++} based on \cref{eq:NonBayesianConfidenceSequence} are shown for reference.
    }
    \label{fig:simulation_noisy_improper}
\end{figure}
\begin{figure}[h]
    \centering
    \includegraphics[width=\textwidth]{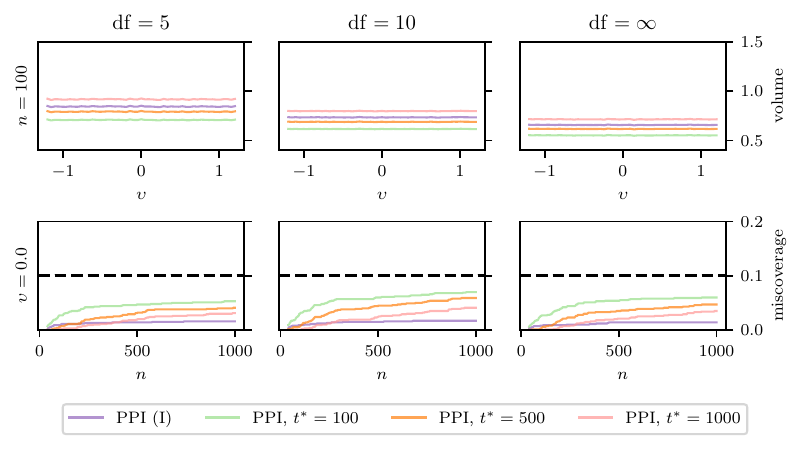}
    \caption{
        Biased predictions study with alternative \AsympCS. The left, middle and right panels show average interval volume and cumulative miscoverage rate over $100$ repetitions for $\mathrm{df} = 5, 10, \infty$. Results for non-assisted PPI based on \cref{eq:NonBayesianConfidenceSequence} are shown for reference.
    }
    \label{fig:simulation_biased_improper}
\end{figure}
\begin{figure}[h]
    \centering
    \includegraphics[width=\textwidth]{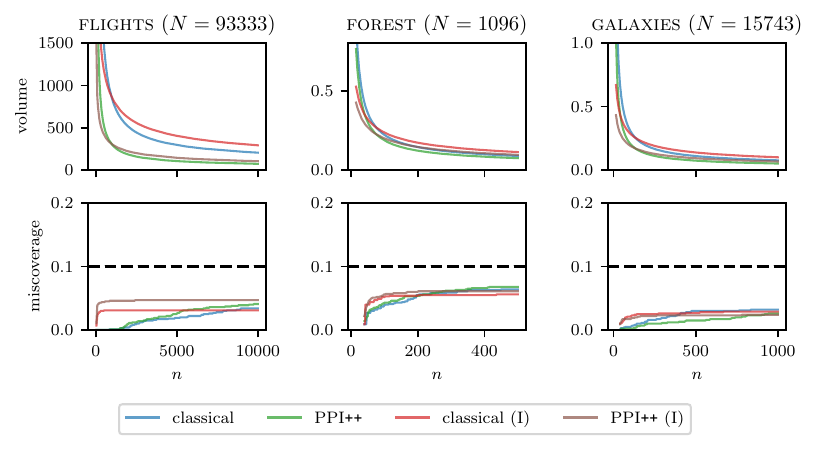}
    \caption{
        Real data study with alternative \AsympCS. The top and bottom rows show the average interval volume and cumulative miscoverage rate over $1000$ repetitions for the \textsc{flights}, \textsc{forest}, and \textsc{galaxies} datasets. Results for classical inference and non-assisted PPI\texttt{++} based on \cref{eq:NonBayesianConfidenceSequence} are shown for reference.
    }
    \label{fig:real_data_improper}
\end{figure}
Compared with the standard non-assisted \AsympCS used in the main text, which depends on the hyperparameter $\rho$, the parameter-free alternative typically performs slightly worse under our default choice of $\rho$ (see \cref{supp:hyperparameters}).
Nonetheless, $\calC^{\NA'}_{\alpha,t}$ can represent an attractive choice when selecting $\rho$ is problematic, precisely because it avoids any tuning.

%\newpage
%\input{Checklist}

\end{document}